\documentclass[10pt,twocolumn,letterpaper]{article}
\usepackage[pagenumbers]{iccv}              

\definecolor{iccvblue}{rgb}{0.21,0.49,0.74}
\usepackage[pagebackref,breaklinks,colorlinks,allcolors=iccvblue]{hyperref}

\usepackage{multirow} 
\usepackage{bbding}
\usepackage{pifont}
\usepackage{wasysym}
\usepackage{amssymb}
\usepackage{bm}
\usepackage{wasysym}

\usepackage{nicematrix}

\usepackage[utf8]{inputenc} %
\usepackage[T1]{fontenc}    %
\usepackage{url}            %
\usepackage{booktabs}       %
\usepackage{amsfonts}       %
\usepackage{nicefrac}       %
\usepackage{microtype}      %

\usepackage{graphicx}
\usepackage{caption}

\usepackage{fontawesome5} %
\usepackage{multirow} 
\usepackage{amsmath} %
\usepackage{xspace} %
\usepackage{titletoc} %
\usepackage{colortbl}
\definecolor{tabhighlight}{HTML}{e5e5e5}
\definecolor{linkcolor}{HTML}{ED1C24}
\definecolor{LGray}{gray}{0.97}

\definecolor{darkgreen}{rgb}{0.0, 0.5, 0.0}

\def\paperID{11454} 
\def\confName{ICCV}
\def\confYear{2025}

\title{Beyond Simple Edits: Composed Video Retrieval with Dense Modifications}

\author{
Omkar Thawakar\textsuperscript{\textnormal{1}}\thanks{Equal Contribution} 
\quad Dmitry Demidov\textsuperscript{\textnormal{1*}}
\quad Ritesh Thawkar\textsuperscript{\textnormal{1}} 
\quad {Rao Muhammad Anwer}\textsuperscript{\textnormal{1}}  \\
\quad  {Mubarak Shah}\textsuperscript{\textnormal{2}}
\quad {Fahad Shahbaz Khan}\textsuperscript{\textnormal{1,3}} 
\quad  {Salman Khan}\textsuperscript{\textnormal{1,4}} \\
\\
\textsuperscript{1}Mohamed bin Zayed University of AI, 
\textsuperscript{2}University of Central Florida, \\
\textsuperscript{3}Linköping University, 
\textsuperscript{4}Australian National University
}

\begin{document}
\maketitle 

\begin{abstract}
Composed video retrieval is a challenging task that strives to retrieve a target video based on a query video and a textual description detailing specific modifications. Standard retrieval frameworks typically struggle to handle the complexity of fine-grained compositional queries and variations in temporal understanding limiting their retrieval ability in the fine-grained setting. 
To address this issue, we introduce a novel dataset that captures both fine-grained and composed actions across diverse video segments, enabling more detailed compositional changes in retrieved video content.
The proposed dataset, named Dense-WebVid-CoVR, consists of 1.6 million samples with dense modification text that is around seven times more than its existing counterpart. We further develop a new model that integrates visual and textual information through Cross-Attention (CA) fusion using grounded text encoder, enabling precise alignment between dense query modifications and target videos. The proposed model achieves state-of-the-art results surpassing existing methods on all metrics. Notably, it achieves 71.3\% Recall@1 in visual+text setting and outperforms the state-of-the-art by 3.4\%, highlighting its efficacy in terms of leveraging detailed video descriptions and dense modification texts. Our proposed dataset, code, and model are available at : \url{https://github.com/OmkarThawakar/BSE-CoVR}.
\end{abstract}

\section{Introduction}
\label{sec:intro}

Composed Image Retrieval (CoIR) aims at retrieving an image from a database based on a reference image and a textual description detailing the desired modifications.  
Recently, Ventura et al.~\cite{webvid-covr} extend this problem to the video domain, giving rise to Composed Video Retrieval (CoVR), where the aim is to retrieve a target video based on a reference video and a modification text. 

Fine-grained CoVR task strives to retrieve a target video based on a reference video and a detailed textual modification, requiring models to capture subtle visual and temporal changes. For example, in video editing and media production, professionals often require systems that can retrieve content with subtle variations in scene composition or actor actions, aiding in locating footage that meets specific creative requirements. As videos grow longer and more complex, users are likely interested in searching for specific moments or actions rather than an entire clip. As a result, fine-grained CoVR task requires a deeper understanding of both the visual and textual information to ensure the modifications described are accurately reflected in the retrieved video.

\begin{figure*}[h!]
    \vspace{-0.1em}
    \centering
    \includegraphics[width=\textwidth]{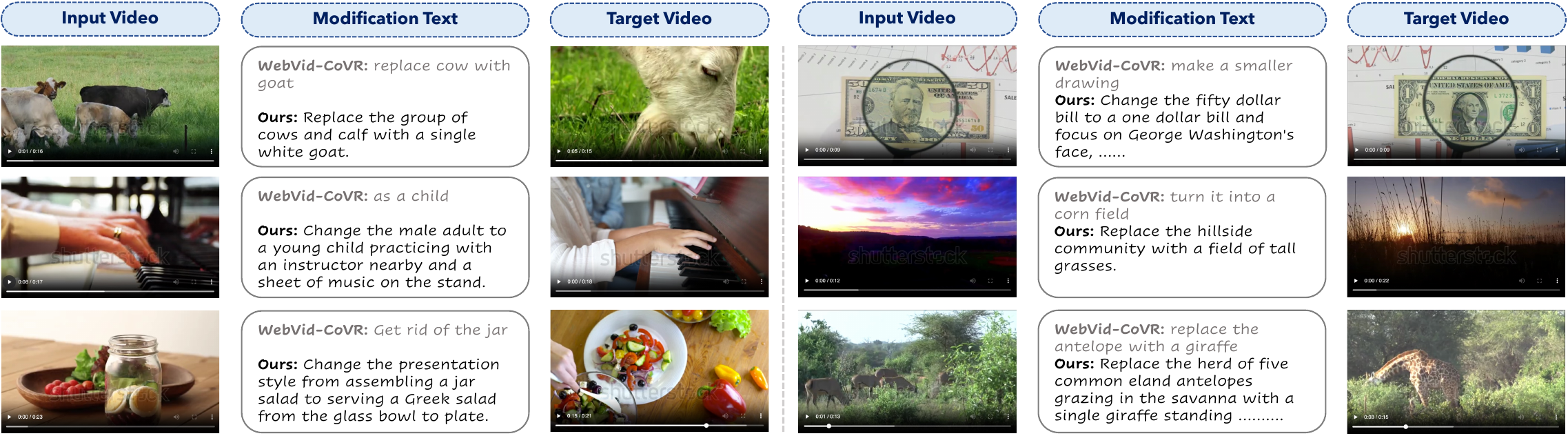}
    \vspace{-2em}
    \caption{
    Example composed video retrieval triplets, consisting of the input video, modification text, and the corresponding target video.
    We compare the basic change text from the existing WebVid-CoVR benchmark~\cite{webvid-covr} with our Dense-WebVid-CoVR dataset that provides a  more detailed and context-aware modification text. Additional examples are in the suppl. material.
    }
    \label{fig:change_text_comparison}
    \vspace{-1em}
\end{figure*}

Fine-grained CoVR task poses unique challenges by requiring the retrieval model to comprehend the visual content as well as the 
intricate temporal sequences and semantic nuances of textual modifications. To this end, existing CoVR benchmarks such as, WebVid-CoVR~\cite{webvid-covr} are insufficient due to lack of granularity, short or generic modification text, and imprecise temporal understanding, which is crucial for capturing subtle visual and temporal modifications (see Fig.~\ref{fig:change_text_comparison}). For instance, in Fig.~\ref{fig:change_text_comparison} example 2, WebVid-CoVR benchmark's difference "as a child" is insufficient to retrieve video with a "young child playing piano with his instructor".

A robust CoVR benchmark must capture subtle yet meaningful differences between input and target videos using context-rich modification texts. While prior benchmarks, such as EgoCVR~\cite{hummel2024egocvr}, focus on specific aspects like temporal changes, they lack dense modification text and dataset diversity, being limited to egocentric videos. These limitations highlight the need for a comprehensive benchmark that effectively encodes visual, semantic, and temporal modifications across a broad range of video content, including lifestyle, nature, sports, and educational domains.

We introduce Dense-WebVid-CoVR, a benchmark designed to enhance retrieval accuracy by leveraging detailed modification texts (see Fig.~\ref{fig:change_text_comparison}). Our dataset is not merely about lengthening modification text, but about providing richer, more contextually grounded descriptions that help retrieval models distinguish subtle yet important changes in videos. Unlike traditional text-to-video retrieval, composed retrieval requires understanding how a target video differs from a reference, making detailed modification texts essential for capturing spatial transformations, object manipulations, and temporal transitions.
Dense-WebVid-CoVR is constructed through a two-step process: (1) Detailed video descriptions are generated using Gemini-Pro~\cite{gemini}, ensuring high-quality, context-aware textual representation. (2) Source-Modification\_Text-Target triplets are then created using GPT-4o~\cite{hurst2024gpt}, which refines video pair relationships with dense, structured modifications. A manual verification step further enhances annotation quality, ensuring that modification texts remain precise, avoid redundancy with the input video, and truly require multimodal understanding.
For example, in Fig.~\ref{fig:change_text_comparison}, our modification text does not simply describe the target video in isolation but explicitly encodes contextual relationships with the input video. In Example 2, rather than a generic phrase like "child playing the piano", our benchmark guides the retrieval system by specifying: "change the male adult to a young child practicing piano with an instructor and a sheet of music on the stand." 
To further mitigate the risk of text-only retrieval overshadowing multimodal learning, our benchmark introduces query structuring strategies that prevent the modification text from being a direct target description. 
Building on our Dense-WebVid-CoVR dataset, we develop a new fine-grained CoVR model that effectively encodes the relationships between visual and textual data by fusing input video, description and dense modification text in a single grounding encoder. Our contributions are summarized as:
\begin{itemize}
    \item We introduce a large-scale fine-grained CoVR dataset, named Dense-WebVid-CoVR, with enriched modification text to encode subtle visual and temporal changes. The dataset comprises 1.6 million samples with an average description length of 81 words and modification text of 31 words, which is around seven times more than the existing CoVR dataset~\cite{webvid-covr}. The test set is fully manually verified to ensure high-annotation quality.
    \item We further develop a robust CoVR model, leveraging our Dense-WebVid-CoVR dataset, that effectively utilizes the rich text modifications together with input video and textual descriptions in a single grounding encoder. On the Dense-WebVid-CoVR test set, our model achieves a gain of 3.4\% over the best existing method \cite{thawakar2024composed} when using the same training set, input modalities, and backbone.

\end{itemize}

\begin{table*}[t!]
    \centering
    \setlength{\tabcolsep}{2pt}
    \resizebox{\textwidth}{!}{
    \begin{tabular}{lcccccccccccc}
        \toprule
        \textbf{Benchmark} & \textbf{Venue} & \textbf{Type} & \textbf{\# Samples} & \textbf{Splits} & \textbf{Data Type} & \textbf{Tools Used} & \textbf{Human} & \textbf{Avg Length of} & \textbf{Avg Length of} & \textbf{Dense} & \textbf{Dense} &\textbf{Diversity}\\
         &  &  &  &  &  &  & \textbf{Verification} & \textbf{Description} & \textbf{Modifiction Text} & \textbf{Modifiction Text} & \textbf{ Descriptions} & \\
        \midrule

        \rowcolor{LGray} InstructPix2Pix & CVPR-2023 & Image & 454K & train, test & synthetic & GPT-3, Stable Diffusion & None & - & 9.4 & \textcolor{red}{\ding{55}} & \textcolor{red}{\ding{55}} & \textcolor{darkgreen}{\checkmark} \\

        \rowcolor{LGray} CIRR & ICCV-2021 & Image & 36K & train, test & real-world & AMT & test & - & 11.3 & \textcolor{red}{\ding{55}} & \textcolor{red}{\ding{55}} & \textcolor{darkgreen}{\checkmark} \\

        \rowcolor{LGray} FashionIQ & ICCV-2019W & Image & 60K & train, test & synthetic & - & - & - & 5.3 & \textcolor{red}{\ding{55}} & \textcolor{red}{\ding{55}} & \textcolor{red}{\ding{55}} \\

        \rowcolor{LGray} CIRCO & ICCV-2023 & Image & 4.5K & test & synthetic & SEARLE & - & - & 8.2 & \textcolor{red}{\ding{55}} & \textcolor{red}{\ding{55}} & \textcolor{darkgreen}{\checkmark} \\

        WebVid-CoVR & AAAI-2024 & Video & 1.6M & train, test & real-world & MTG-LLM & test & 6.68 & 4.6 & \textcolor{red}{\ding{55}} & \textcolor{red}{\ding{55}} & \textcolor{darkgreen}{\checkmark} \\


        Ego-CVR & ECCV-2024 & Video & 2.2K & test & real-world & GPT-4 & test & - & 4 & \textcolor{red}{\ding{55}} & \textcolor{red}{\ding{55}} & \textcolor{red}{\ding{55}} \\
        

        \rowcolor{orange!10}\textbf{Dense-WebVid-CoV (Ours)} & - & Video & 1.6M & train, test & real-world & Gemini-Pro, GPT-4 & train, test & \textbf{81.32} & \textbf{31.16} & \textcolor{darkgreen}{\checkmark} & \textcolor{darkgreen}{\checkmark} & \textcolor{darkgreen}{\checkmark} \\
        
        \bottomrule
    \end{tabular}
    }
    \vspace{-1em}
    \caption{\textbf{Comparative analysis of various video and image-based benchmarks for composed video retrieval (CoVR)}. The benchmarks are categorized by type (video or image), sample size, data type (real-world or synthetic), tools used for generation, and human verification. Compared to existing CoVR datasets, our benchmark provides fine-grained dense descriptions with an average description length of 81.32 words and modification texts of 31.16 words, surpassing them in generating rich, context-aware video retrieval capabilities.
    }
    \label{tab:benchmark_compare}
\vspace{-1em}
\end{table*}

\section{Related Work}

\noindent\textbf{Composed Image Retrieval (CoIR):}
%
 The task aims to retrieve images based on a reference image and a modification text describing desired changes \cite{vo2019composing}. 
Earlier methods rely on manually annotated datasets, e.g., CIRR~\cite{cirr} and FashionIQ~\cite{fashioniq}, which are of high-quality but limited in scale due to the labor-intensive annotation process. 
More recent approaches aim to scale the task by automatically generating large datasets. 
Recently, large-scale datasets like LaSCo~\cite{levy2023data} and SynthTriplets18M~\cite{gu2023compodiff} have been generated automatically, using visual question answering and text-conditioned image editing frameworks \cite{brooks2022instructpix2pix}. However, these datasets are not yet publicly available. \\
\noindent\textbf{Composed Video Retrieval (CoVR):}
The CoVR task aims to retrieve target videos based on reference videos and textual modification prompts.
Recent CoVR methods adapt CoIR techniques to video domain by aggregating multi-frame features~\cite{rasheed2023fine,xu2021videoclip,xue2022clip,yang2021taco}. 
Large datasets like WebVid-CoVR~\cite{webvid-covr} and vision-language models (VLMs)~\cite{li2023blip,radford2021learning} have advanced CoVR using contrastive learning and multi-modal embeddings. WebVid-CoVR~\cite{webvid-covr} is constructed by mining video-pairs with similar captions and generating modification texts using large language models (LLMs). Recently, \cite{thawakar2024composed} proposes an approach that utilizes language descriptions of source to improve query-specific alignment between the source and target videos. The work of~\cite{hummel2024egocvr} introduces an action retrieval benchmark from egocentric videos. However, the benchmark only has a test set, does not capture dense modification text and is limited to egocentric videos. The benchmark also lacks diversity in terms of video content (e.g., general purpose videos including, nature, sports, educational visual content and lifestyle), that is available in WebVid-CoVR dataset. In this work, we develop a new benchmark (see Tab.~\ref{tab:benchmark_compare}) that comprises both training and test set, while containing high-quality detailed modification text about general purpose videos.

\section{Dense-WebVid-CoVR Benchmark}


To construct a fine-grained CoVR benchmark with detailed modification text, we start with the WebVid-CoVR dataset~\cite{webvid-covr} that contains diverse video content including nature, lifestyle, and professional activities. The dataset contains 1.6 million triplets with videos averaging around 16.8 seconds in length. To effectively create captions for videos, we employ the Gemini-Pro~\cite{gemini} model for video captioning. To ensure caption quality, we apply a hallucination check using the BLIP model~\cite{li2023blip} to compute the cosine similarity between the video and its caption. Here, captions scoring below a specified threshold that is set empirically are deemed inadequate and recomputed, ensuring alignment with video content and creating a more robust and reliable dataset. Additional details are presented in suppl. material.

\begin{figure*}[t!]
    \centering
    \includegraphics[width=1\textwidth]{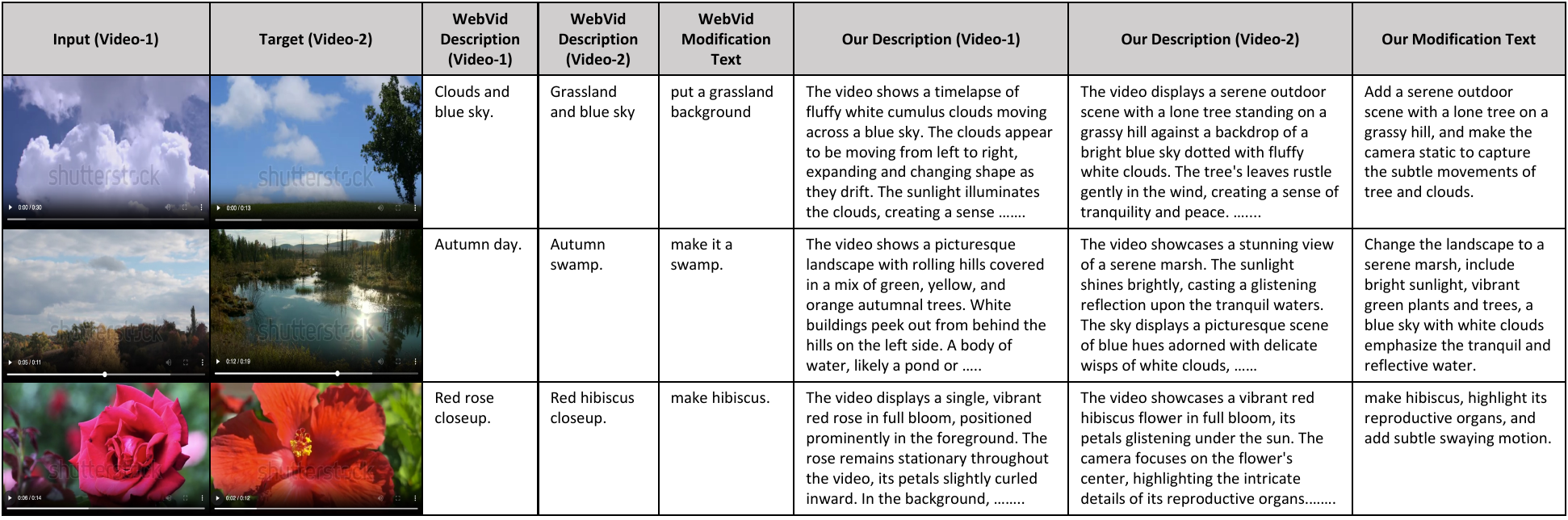}
    \vspace{-2em}
    \caption{Comparison between original WebVid-CoVR~\cite{webvid-covr} descriptions and change-text vs. our generated detailed descriptions and change text. Each row presents an input video (Video-1), a target video (Video-2), and their corresponding descriptions followed by the change-text generated using the descriptions. The original WebVid-CoVR's~\cite{webvid-covr} change texts lack fine-grained details, whereas our approach offers significantly more comprehensive and context-rich change texts. 
    }
    \label{fig:dataset_compare}
    \vspace{-1.2em}
\end{figure*}

\subsection{Modification-Text Generation}

In the context of fine-grained robust CoVR, modification text is crucial and bridges the gap between two similar videos by explicitly describing the differences between them. These descriptions highlight specific changes, such as alterations in actions, objects, or scenes. Dense modifications are expected to cover nuanced variations in visual content to enable more precise retrieval of videos based on subtle details. For this, we employ GPT-4o~\cite{hurst2024gpt} to generate modification texts. 
We provide the model with existing triplets from WebVid, including both the original video captions and their corresponding modification texts, to guide the generation process.
Fig.~\ref{fig:dataset_compare} shows the differences between WebVid descriptions and our dense descriptions and detailed modification text. Compared to original WebVid descriptions, subtle changes between videos are captured. For instance Fig.~\ref{fig:dataset_compare} row 1 contains a outdoor nature video. The original WebVid modification text (\textit{put a
grassland background}) lacks subtle details. In contrast, our modification text (\textit{Add a serene outdoor scene with a lone tree on a
grassy hill, and make the camera static to capture the subtle movements of the tree and clouds, evoking tranquility and the
beauty of nature better captures}) better captures detailed information. The examples in Fig.~\ref{fig:dataset_compare} shows that our modification text provides more details in the form of intricate visual elements, such as environmental settings, colors, lighting, and subtle changes in object focus.

\subsection{Modification-Text Verification}



To ensure high-quality, we manually verify the generated modification texts. During this process, input and target videos are presented side-by-side along with their generated dense descriptions. The annotators are tasked with assessing the quality of the modification text by comparing it to the visual content and making corrections when necessary. 
%
%
To ensure the quality and accuracy of the generated modification text, the following quality control protocol is used for verification process. (i) A side-by-side comparison to ensure that the changes mentioned accurately reflect the differences between the two videos. (ii) A contextual consistency check to verify that the modification text addresses key changes, such as consistent object movement and transitions in the main scene, related surroundings, and background. (iii) An action and object verification check, where the annotators are asked to check whether the objects and their corresponding actions mentioned in the modification text are present in both the videos. (iv) A temporal alignment check to ensure that the modification text aligns with the actual sequence of actions in the videos. (v) The annotators are asked to check the comprehensive description quality to ensure all relevant changes between the input and target video are covered. (vi) The annotators are asked to ensure that the modification text is clear and concise. (vii) We  empirically set  a cosine similarity threshold  to identify low-quality modification texts for further manual  modification. Moreover, it is worth noting that annotators are asked to make manual corrections to the modification text in case of any errors or missing details.

We manually verify \textit{all} 3,000 triplets each from the test set and validation set consisting of WebVid-8M~\cite{webvid2m} corpus videos. For training set, out of the total of 1.6M triplets consisting WebVid-2M~\cite{webvid2m} corpus, we carefully select 100k triplets with unique input and target video covering distinct categories representing all training triplets for verification. 
To further prevent text-only retrieval from overshadowing multimodal learning, we introduce query structuring strategies that ensure modification texts are not direct target descriptions but require contextual interpretation with the input video. This enforces true multimodal reasoning, ensuring that models cannot retrieve the target video based solely on modification text as stated in CIRCO~\cite{agnolucci2024isearle}. These structuring strategies play a crucial role in preserving the core purpose of Composed Video Retrieval (CoVR) by preventing the task from being reduced to text-to-video retrieval.
To ensure high-quality annotations, trained annotators manually verified and refined modification texts through multiple rounds. Although around 2-3\% of modification texts in the training set may have minor inaccuracies, our experiments show that this has minimal impact on model performance. Instead, the inclusion of detailed modification texts significantly improves retrieval accuracy, leading to a 3.4\% gain in Recall@1, proving that our dataset remains highly reliable and effective for fine-grained video retrieval. Additional details are presented in the suppl. material (Section E). 
Next, we introduce our method that leverages the detailed modification text from the  Dense-WebVid-CoVR dataset for fine-grained CoVR.

\section{Method}

\textbf{Problem Formulation:} In the fine-grained CoVR task, the objective is to retrieve a modified video from a large database using two inputs: a reference video and a detailed textual description that outlines the desired modification. Let $\mathcal{V}$ represent the set of all videos, $\mathcal{D}$ represent the set of corresponding dense descriptions, and $\mathcal{T}$ the space of detailed textual modifications. Given a query video $q \in \mathcal{V}$, its description $d \in \mathcal{D}$ and a corresponding detailed textual modification $t \in \mathcal{T}$, the goal is to identify the target video $v^* \in \mathcal{V}$ that best reflects the described changes. The retrieval system is desired to leverage both visual and textual embeddings to capture semantic changes, ensuring fine-grained, context-sensitive video retrieval. 

\begin{figure*}[t!]
    \centering
    \includegraphics[width=1\textwidth]{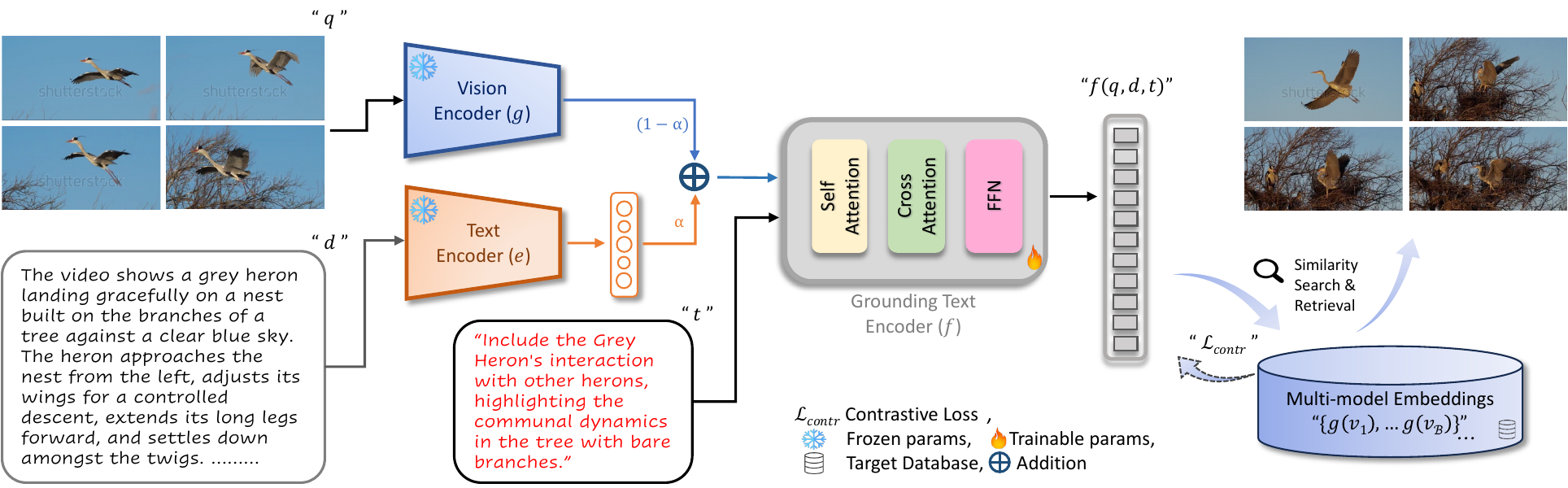}
    \vspace{-2em}
    \caption{
    Our proposed fine-grained CoVR architecture comprising three main components: a vision encoder ($g$), a text encoder ($e$), and a grounding text encoder ($f$). The video query ($q$) and detailed textual description ($d$) are processed by vision encoder ($g$) and text encoder ($e$), with a projection layer aligning their embeddings. 
    Through a unified fusion strategy, the weighted combination of these embeddings is grounded with the modification text ($t$) in the grounding text encoder ($f$) using self-attention, cross-attention, and a feed-forward network (FFN).
    The model is trained with contrastive loss ($\mathcal{L}_{contr}$), leveraging frozen ($g,e$) and trainable ($f$) components to produce multi-modal embeddings for effective video retrieval.
    }
    \label{fig:architecture}
    \vspace{-1.2em}
\end{figure*}

\begin{figure}[t]
    \centering
    \includegraphics[width=1\linewidth]{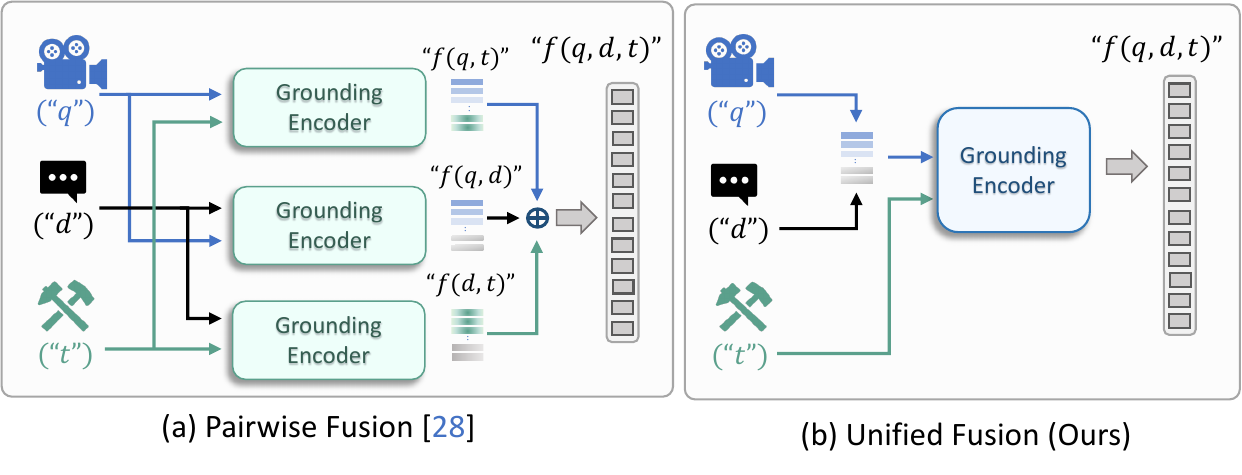}
    \vspace{-2em}
    \caption{Comparison between the pairwise fusion strategy~\cite{thawakar2024composed} (left) and our unified fusion scheme (right) for integrating query ($q$), description ($d$), and modification text ($t$) in CoVR. 
    }
    \label{fig:fusion_compare}
    \vspace{-1.5em}
\end{figure}

\subsection{Overall Architecture}

For the fine-grained CoVR task, it is desired to accurately encode fine-grained subtle visual and temporal changes by capturing the inter-dependencies between the query video ($q$), dense detailed description ($d$), and the modification text ($t$). Recent methods, such as ~\cite{thawakar2024composed} utilizes a pairwise fusion scheme (see Fig.~\ref{fig:fusion_compare}a) that processes each component pair separately (e.g., $f(q, t)$, $f(d, t)$, and $f(q, d)$). We observe such strategy to achieve sub-optimal results likely due to the dilution of the rich modification text details in the final multimodal embedding. To this end, we introduce an approach having a simple yet effective fusion strategy to better align the multimodal query and target videos.
\noindent Our architecture (see Fig.~\ref{fig:architecture}) comprises three components: vision encoder ($g$), text encoder ($e$), and grounding text encoder ($f$). The goal is to retrieve a target video based on a query video ($q$), a detailed description of the query ($d$), and a textual modification ($t$) that describes the desired changes.

\noindent\textbf{Vision Encoder:} The Vision Encoder ($g$) utilizes ViT-L~\cite{dosovitskiy2020image} as its backbone and processes the visual input. Instead of processing every frame, the middle frame of the input video is selected to compute the visual embeddings for efficient feature extraction, following~\cite{webvid-covr,thawakar2024composed}. 

\noindent\textbf{Text Encoder:} The text encoder ($e$), pretrained from the BLIP, processes the detailed textual description ($d$) that accompanies each video. The comprehensive description is expected to comprise both spatial aspects and temporal elements to effectively summarize all actions within the video. Consequently, the description embeddings capture all video-level features to obtain a holistic representation of the input video. To obtain alignment between visual and textual features, a projection layer is employed that aligns the text embeddings with the visual embeddings extracted by the vision encoder ($g$). These aligned embeddings are then integrated using a learnable parameter $\alpha$, as in:
\begin{equation}
\small
{embs} = (1 - \alpha)g(q) + \alpha e(d)
\end{equation}
This enables the model to dynamically balance the impact of visual and textual information (see Fig.~\ref{fig:architecture}) based on the optimized $\alpha$ value derived from validation set.  


\noindent\textbf{Grounding Text Encoder:} The grounding text encoder ($f$) generates the final composed multimodal embedding. The  encoder $f$ fuses the query video and description embeddings with the detailed modification text ($t$) using a cross-attention mechanism. Cross-attention layers align the visual features from the video with the textual description, grounding the modification in the correct context. 
The encoder $f$ outputs a fused embedding ($f(q,d,t)$) that encodes the query, description, and modification text in a single representation. This enriched embedding is then compared against embeddings of target videos in a large video database for retrieval.

\begin{figure}[t]
    \centering
    \includegraphics[width=1\linewidth]{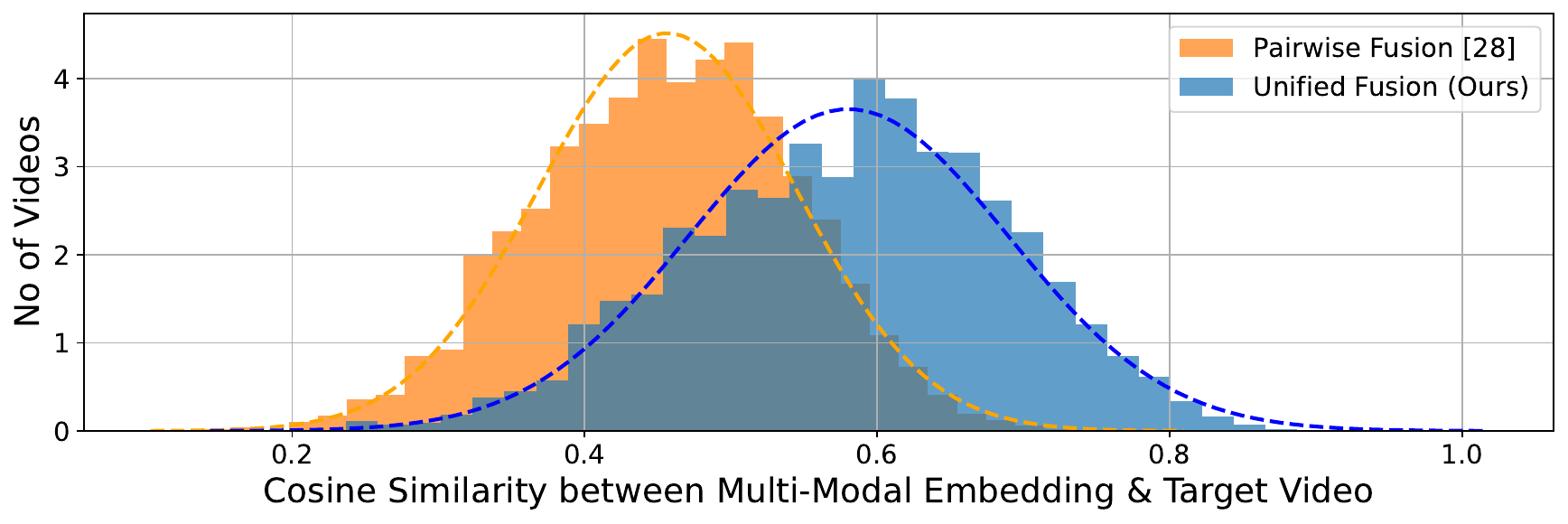}
    \vspace{-2em}
    \caption{ Comparison between the pairwise fusion~\cite{thawakar2024composed}  and our unified fusion in terms of cosine similarity score computed between multimodal query embedding and target video. The results are computed on Dense-WebVid-CoVR validation set. The pairwise embeddings are generated separately, leading to less contextually aligned representations as each pairwise combination is processed separately. In contrast, our method fuses query ($q$) and description ($d$) first followed by combination with modification text($t$) within a \textit{single} grounding encoder. 
    Our simple yet effective fusion scheme enables richer and granular understanding of the relationships between visual and textual data achieving higher similarity between multi-modal embeddings with target videos.  
    }
    \label{fig:method_compare}
    \vspace{-1.7em}
\end{figure}

\begin{table*}[t!]
    \centering
    \setlength{\tabcolsep}{8pt}
    \resizebox{\textwidth}{!}{%
    \begin{tabular}{clclccc|cccc}
        \toprule
        & & Training  &  & Modification &  &   &  \multicolumn{4}{c}{Recall@K}  \\
        & Model &  Dense-WebVid-CoVR & Input Modalities & Text Fusion & Backbone &  Frames & R@1 & R@5 & R@10 & R@50 \\
        \midrule

        1 & Random & & - & - & - & - & 0.04 & 0.21 & 0.32 & 1.46 \\
        2 & CoVR-BLIP \cite{webvid-covr}  & \ding{56} & Text & - & BLIP & - & 24.12 & 56.02 & 60.16 & 82.34 \\
        3 & CoVR-BLIP \cite{webvid-covr}  & \ding{56} & Visual & - & BLIP & 15 & 22.52 & 53.08 & 58.34 & 81.26 \\
        4 & CoVR-BLIP \cite{webvid-covr}  & \ding{56} & Visual + Text & Avg & BLIP & 15 & 38.44 & 64.96 & 71.72 & 87.12 \\
        5 & Thawakar .\textit{et.al.} \cite{thawakar2024composed} & \ding{56} & Visual + Text & Avg & BLIP & 15 & 40.23 & 66.38 & 74.84 & 88.12 \\
        \rowcolor{orange!5}
        6 & \textbf{Our Approach} & \ding{56}  & Visual + Text & Avg & BLIP & 15 & \textbf{42.41} & \textbf{68.54} & \textbf{77.07} & \textbf{91.24} \\
        7 & CoVR-BLIP \cite{webvid-covr} & \ding{56} & Visual + Text & CA & BLIP & 15 & 35.60 & 60.80 & 70.31 & 87.05 \\
        8 & Thawakar .\textit{et.al.} \cite{thawakar2024composed} & \ding{56} & Visual + Text & CA & BLIP & 15 & 39.20 & 64.40 & 75.56 & 88.90 \\
        \rowcolor{orange!10}
        9 & \textbf{Our Approach} & \ding{56}  & Visual + Text & CA & BLIP & 15 & \textbf{48.08} & \textbf{73.36} & \textbf{81.06} & \textbf{93.78} \\
        \midrule
        10 & CoVR-BLIP \cite{webvid-covr}  & \ding{52}  & Text & - & BLIP & - & 38.92 & 66.36 & 74.08 & 90.92 \\
        \rowcolor{orange!5}
        11 & \textbf{Our Approach} & \ding{52} & Text & - & BLIP & 15 & \textbf{50.12} & \textbf{78.36} & \textbf{79.52} & \textbf{92.62} \\
        12 & CoVR-BLIP \cite{webvid-covr} & \ding{52}  & Visual & - & BLIP & 15 & 36.26 & 64.32 & 72.18 & 90.46 \\
        13 & CoVR-BLIP \cite{webvid-covr} & \ding{52}  & Visual + Text & CA & BLIP & 15 & 63.12 & 85.66 & 92.56 & 97.36 \\
        14 & Thawakar .\textit{et.al.} \cite{thawakar2024composed} & \ding{52}  & Visual + Text & CA & BLIP & 15 & 67.86 & 87.72 & 93.06 & 98.18 \\
        \rowcolor{orange!10}
        15 & \textbf{Our Approach} & \ding{52} & Visual + Text & CA & BLIP & 15 & \textbf{71.26} & \textbf{89.12} & \textbf{94.56} & \textbf{98.88} \\
        \bottomrule
    \end{tabular}
    }
    \vspace{-0.8em}
    \caption{\textbf{Comparison of our approach with existing methods on the Dense-WebVid-CoVR test set}. Our proposed approach consistently outperforms existing methods in \textit{all} settings and Recall@K metrics. 
    Notably in the Visual + Text setting with Cross-Attention (CA), our method improves Recall@1 to 71.26 and Recall@50 to 98.88. Best results are in bold.}
    \label{tab:sota_comparison}
    \vspace{-1em}
\end{table*}

As discussed above, our approach employs a unified fusion scheme to simultaneously fuse $q$, $d$, and $t$ by integrating all three components within a \textit{single} grounding encoder  (see Fig.~\ref{fig:fusion_compare}b). Compared to the pairwise fusion~\cite{thawakar2024composed}, our unified fusion achieves better alignment between multi-modal query and target videos (see Fig.~\ref{fig:method_compare}). 
The overall model is trained using a contrastive loss ($\mathcal{L}_{contr}$), encouraging the alignment of fused embeddings ($f(q,d,t)$) with correct target video embeddings ($g(v)$). The objective is defined as minimizing:

\begin{align}
\scriptsize
\label{eq:hn_nce}
\mathcal{L}_{contr}= -\sum_{i \in \mathcal{B}}\text{log}\left(\frac{e^{S_{i,i}/\tau}}{\lambda \cdot {e^{S_{i,i}/\tau}} + \sum_{j \neq i}{e^{S_{i,j}/\tau}w_{i,j}} }\right) \nonumber \\\scriptsize - \sum_{i \in \mathcal{B}} \text{log}\left(\frac{e^{S_{i,i}/\tau}}{\lambda \cdot {e^{S_{i,i}/\tau}} + \sum_{j \neq i}{e^{S_{j,i}/\tau}w_{j,i}} }\right)
\end{align}
Here, $\lambda$ is assigned a value of 1, and the temperature $\tau$ is set to 0.07, following~\cite{hn_nce}. The term $S_{i,j}$ represents the cosine similarity between the joint composed multi-modal embedding ${f}(q,d,t)$ and the corresponding target video $g(v)$. The weight $w_{i,j}$ is configured as in \cite{hn_nce}, using $\beta=0.5$, and $\mathcal{B}$ refers to batch size.
Finally, a similarity search is performed over video database using the fused multi-modal embeddings. Here, we compare the embedding ($f(q,d,t)$) with embeddings of all target videos ($g(v)$). Videos with highest similarity scores are retrieved as the most likely matches. 


\section{Experiment}

\textbf{Datasets: } We conduct experiments on the proposed Dense-WebVid-CoVR dataset. The training set comprises 131K distinct videos paired with 467K unique change texts, having video descriptions with average words length 81.32 with each video associated with an average of 12.7 triplets, and the change texts averaging 31.2 words in length. The dataset also includes 7K validation triplets and 3.2K manually curated test triplets, ensuring high-quality evaluation sets from the WebVid10M corpus. In addition to Dense-WebVid-CoVR, we conduct experiments on the recent EgoCVR~\cite{hummel2024egocvr} to further evaluate our approach on Ego-Centric videos. EgoCVR consists of 2,295 samples, with 78.9\% focusing on temporal events and 21.1\% on object-centered modifications, focusing on temporal video understanding. 

\begin{figure*}[t]
    \centering
    \includegraphics[width=1\linewidth]{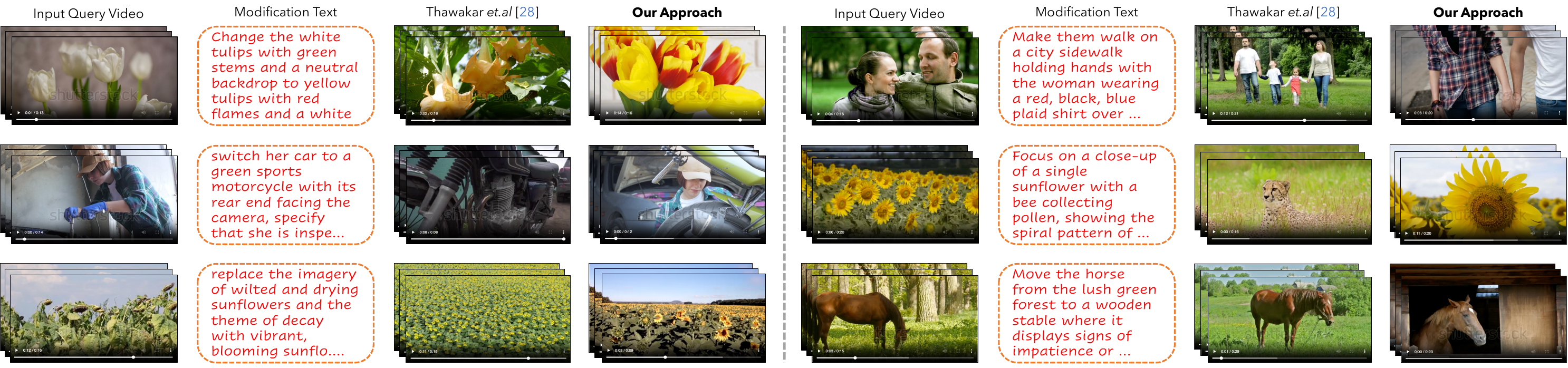}
    \vspace{-2em}
    \caption{
   Qualitative comparison between the recent CoVR method~\cite{thawakar2024composed} and our approach on example videos from the  
    Dense-WebVid-CoVR test set. 
    The approach of~\cite{thawakar2024composed} based on pairwise fusion misses fine-grained details from the modification text, leading to sub-optimal retrieval performance. For instance in the first example video (row 1 on the left), it retrieves the video with yellow flower but misses the other details of \textit{red flames and a white brick background} in the modification text. Similarly, it misses the fine-grained details in the modification text such as, \textit{specify that she is inspecting ...} in the first example (left) on row 2. Our approach 
    achieves superior retrieval performance by better capturing the fine-grained details and context. Best viewed zoomed in. Additional results are presented in suppl. material.
    }
    \label{fig:qual_comparison}
    \vspace{-1em}
\end{figure*}

We also conduct experiments on composed image retrieval (CoIR) task using two standard benchmarks: CIRR~\cite{cirr} and FashionIQ~\cite{fashioniq}. CIRR contains 36.5K manually annotated open-domain natural image pairs along with their change text. The dataset is split into 28.2K, 16.7K pairs for training, and 41.8K, 22.6K pairs for testing and validation, respectively. FashionIQ focuses on fashion products in three categories such as Shirts, Dresses, and Tops/Tees consisting of 30K images paired with 40.5K change texts. The data distribution includes 18K, 45.5K pairs for training, and 60.2K, 15.4K for testing and validation. 


\noindent\textbf{Evaluation Metrics:} We follow standard protocols for both composed video and image retrieval tasks (CoVR and CoIR), as in prior works~\cite{cirr,webvid-covr,hummel2024egocvr}. Retrieval performance is measured using Recall@K (R@k), where k represents the top-k ranked results. Recall at rank k signifies the percentage of times the correct target is retrieved within the top-k results. Specifically, we report the recall values at ranks 1, 5, 10, and 50 to comprehensively assess the model's retrieval accuracy across different scenarios. \\
\noindent\textbf{Implementation Details: } 
We utilize ViT-L~\cite{dosovitskiy2020image} as the vision encoder. The text encoder and the grounding text encoder is from BLIP-2~\cite{li2023blip}, for fusing the modification-text with multimodal features. The model is trained for 5 epochs with a batch size of 1024 (256 per device) and an initial learning rate of $1e-5$. The value of learnable parameter $\alpha$ derived from validation set is $0.36$.  For transfer learning on CoIR, we fine-tune on the FashionIQ dataset for 6 epochs using a batch size of 2048/1024 and a learning rate of $1e-4$. All experiments are run on four NVIDIA A100 40GB GPU's.

\subsection{Results on Composed Video Retrieval (CoVR)}

Tab.~\ref{tab:sota_comparison} presents a comparison with existing methods on the proposed Dense-WebVid-CoVR test set with dense modification text under different settings. Our approach achieves consistently improved performance in all settings and metrics. 
When using the training setting and both input modalities (visual + text), the recent work of 
Thawakar .\textit{et.al.} \cite{thawakar2024composed} obtains Recall@1 and Recall@5 score of 67.9 and 87.7,respectively. Our approach outperforms \cite{thawakar2024composed} with  Recall@1 and Recall@5 score of 71.3 and 81.1, respectively. 
In case of no training scenario and cross-attention (CA) based modification text fusion, our approach achieves a larger performance (+8.9\% in Recall@1) over ~\cite{thawakar2024composed} likely due to the proposed unified fusion being more effective at leveraging the detailed descriptions. Additionally, Our approach shows impressive performance of Recall@1 of 50.12 with text-based video retrieval.
We further note that our approach is 3$\times$ times faster than~\cite{thawakar2024composed} due to the unified fusion approach that avoids repetition (in contrast to pairwise fusion) and utilizes a single grounding text encoder to construct multimodal embedding.

Fig.~\ref{fig:qual_comparison} presents a qualitative between~\cite{thawakar2024composed} employing pairwise fusion and our approach based on unified fusion on example videos from Dense-WebVid-CoVR test set. We observe that pairwise fusion-based method~\cite{thawakar2024composed} that separately process each input component misses fine-grained details within the modification text, leading to sub-optimal retrieval quality. Our approach integrating all input elements within a single grounding encoder better captures the context and fine-grained details within the modification text, leading to superior retrieval performance.

\begin{table}[t!]

    \centering
    \resizebox{\linewidth}{!}{%
    \centering
    \setlength{\tabcolsep}{8pt}
    \begin{tabular}{lcccccc}
    \toprule
    \textbf{Method} & \multicolumn{3}{c}{\textbf{Global}} & \multicolumn{3}{c}{\textbf{Local}} \\
    \cmidrule(lr){2-4} \cmidrule(lr){5-7}
     & R@1 & R@5 & R@10 & R@1 & R@2 & R@3 \\
    \midrule
    Random & 0.01 & 0.05 & 0.1 & 25.3 & 38.2 & 50.7 \\
    \midrule
    CoVR-BLIP \cite{webvid-covr} & 5.4 & 15.2 & 24.3 & 33.1 & 49.5 & 62.9 \\
    Thawakar \textit{et.al} \cite{thawakar2024composed} & 6.0 & 14.8 & 24.3 & 33.4 & 49.3 & 63.0 \\
    CIReVL \cite{karthik2023vision} & 2.0 & 6.8 & 10.2 & 21.6 & 35.1 & 46.0 \\
    TFR-CVR~\cite{hummel2024egocvr}  & 14.1 & 39.5 & 54.4 & 44.2 & 61.0 & 73.2 \\
    \rowcolor{orange!10}
    \textbf{Our Approach} & \textbf{14.6} & \textbf{41.3} & \textbf{54.9} & \textbf{44.8} & \textbf{61.7} & \textbf{74.0} \\
    \bottomrule
    \end{tabular}
    }
    \vspace{-0.5em}
    \caption{Comparison of our method with existing approaches, in a zero-shot setting, on the Ego-CVR test set. Our approach performs favorably in terms of both global and local retrieval metrics (R@K) compared to existing methods. Best results are in bold.
    }
    \label{tab:egocvr_comparison}
    \vspace{-1em}
    
\end{table}

\begin{table}[t]
    \centering
    \setlength{\tabcolsep}{5pt}
    \resizebox{0.48\textwidth}{!}{%
    \begin{tabular}{l|c|c|c|c}
        \toprule
        \textbf{Video Descriptions ($d$)} & \textbf{Modification Text ($t$)} & \textbf{R@1} & \textbf{R@5} & \textbf{R@10} \\
        \midrule
        WebVid-CoVR & WebVid-CoVR & 60.4 & 84.5 & 91.4 \\
        WebVid-CoVR & Dense-WebVid-CoVR & 61.2 & 84.8 & 92.6 \\
        Dense-WebVid-CoVR & WebVid-CoVR & 63.8 & 87.5 & 92.4 \\
        \rowcolor{orange!10}
        Dense-WebVid-CoVR & Dense-WebVid-CoVR & \textbf{71.2} & \textbf{89.1} & \textbf{94.5 } \\
        \bottomrule
    \end{tabular}
    }
    \vspace{-0.7em}
    \caption{Ablation study comparing the performance of our proposed model trained with different Video Descriptions ($d$) and Modification Text ($t$) combinations. We report Recall@K scores.}
    \label{tab:ablation_study2}
\vspace{-1em}
\end{table}

\begin{table*}[t!]
    \centering
    \setlength{\tabcolsep}{5pt}
    \begin{minipage}{0.48\textwidth}
        \resizebox{\textwidth}{!}{%
        \begin{tabular}{clc|ccc|cc}
        \toprule
         & & Pretrain  & \multicolumn{3}{c|}{Recall@K} & \multicolumn{2}{c}{$\text{R}_{\text{subset}}$@K} \\
         & Method & Data & K=1 & K=10 & K=50 & K=1  & K=3 \\ 
        \midrule
        \multirow{12.2}{*}{
        \begin{tabular}[c]{@{}c@{}}
        \rotatebox{90}{\small{Train CIRR}}\end{tabular}}
         & TIRG \citep{tirg} & - & 14.61  & 64.08 & 90.03 & 22.67  & 65.14 \\
         & MAAF-RP \citep{MAAF} & - & 10.22  & 48.68 & 81.84 & 21.41  & 61.60 \\
         & ARTEMIS \citep{ARTEMIS} & - & 16.96  & 61.31 & 87.73 & 39.99  & 75.67 \\
         & CIRPLANT \citep{cirr} & - & 19.55  & 68.39 & 92.38 & 39.20  & 79.49 \\
         & LF-BLIP \citep{cclip} & - & 20.89  & 61.16 & 83.71 & 50.22  & 86.82 \\
         & CompoDiff~\citep{gu2023compodiff} & \ding{51} & 22.35  & 73.41 & 91.77 & 35.84  & 76.60 \\ %
         & Combiner \citep{cclip} & - & 33.59  & 77.35 & 95.21 & 62.39  & 92.02 \\
         & CASE~\citep{levy2023case} & \ding{51} & {49.35}  & 88.75 & 97.47 & 76.48  & 95.71 \\ 
         & CoVR-BLIP ~\citep{webvid-covr}  & - & 48.84  & 86.10 & 94.19 & 75.78 & 92.80 \\
         & CoVR-BLIP ~\citep{webvid-covr}  & \ding{51} & 49.69  & 86.77 & 94.31 & 75.01 & 93.16 \\
         & Thawakar \textit{et.al} ~\citep{thawakar2024composed} & \ding{51}  & 51.03 & 88.93 & 97.53 & 76.51  & 95.76 \\
        \rowcolor{orange!15} \cellcolor{white}
         & \textbf{Our Approach} & \ding{51} & \textbf{56.30} & \textbf{91.84} & \textbf{98.20} & \textbf{79.16} & \textbf{96.42} \\ 
        \midrule
        \multirow{9.2}{*}{\begin{tabular}[c]{@{}c@{}} 
        \rotatebox{90}{\small{Zero Shot}}\end{tabular}} 
         & Random$\dagger$  & - & \textcolor{white}{0}0.04 & \textcolor{white}{0}0.44 & \textcolor{white}{0}2.18 & 16.67 & 50.00 \\
         & CompoDiff \citep{gu2023compodiff} & \ding{51} & 19.37 & 72.02 & 90.85 & 28.96 & 67.03 \\ %
         & Pic2Word~\citep{saito2023pic2word} & \ding{51} & 23.90  & 65.30 & 87.80 & - & - \\
         & CASE~\citep{levy2023case} & \ding{51} & 35.40  & 78.53 & 94.63 & 64.29  & 91.61 \\ 
         & CoVR-BLIP ~\citep{webvid-covr} & \ding{51}  & 38.48 & 77.25 & 91.47 & 69.28  & 91.11 \\
         & Thawakar \textit{et.al} ~\citep{thawakar2024composed} & -  & 21.34 & 52.37 & 74.92 & 64.66  & 90.87 \\
         & \textbf{Our Approach} & - & \textbf{32.16} & \textbf{63.34}  & \textbf{78.92} & \textbf{68.62}  & \textbf{91.06} \\
         & Thawakar \textit{et.al} ~\citep{thawakar2024composed} & \ding{51}  & 40.12 & 78.86 & 94.69 & 70.47  & 92.12 \\
        \rowcolor{orange!15}\cellcolor{white}
         & \textbf{Our Approach} & \ding{51} & \textbf{44.08} & \textbf{81.72}  & \textbf{95.88} & \textbf{74.12}  & \textbf{93.18} \\ 
        \bottomrule
        \end{tabular}
        }
    \end{minipage}
    \hfill
    \begin{minipage}{0.48\textwidth}
        \resizebox{\textwidth}{!}{%
        \begin{tabular}{clc|cc|cc|cc}
        \toprule
        & & Pretrain & \multicolumn{2}{c}{Dress} & \multicolumn{2}{c}{Shirt} & \multicolumn{2}{c}{Toptee} \\
        & Method & Data & R@10 & R@50 & R@10 & R@50 & R@10 & R@50  \\
        \midrule
        \multirow{18.2}{*}{
        \begin{tabular}[c]{@{}c@{}}
        \rotatebox{90}{\small{Train FashionIQ}} \end{tabular}}
        & JVSM \citep{JVSM} & - & 10.70 & 25.90 & 12.00 & 27.10 & 13.00 & 26.90  \\
        & CIRPLANT \citep{cirr} & - & 17.45 & 40.41 & 17.53 & 38.81 & 61.64 & 45.38  \\
        & TRACE \citep{TRACE} & - & 22.70 & 44.91 & 20.80 & 40.80 & 24.22 & 49.80  \\
        & VAL w/GloVe \citep{VAL_IR} & - & 22.53 & 44.00 & 22.38 & 44.15 & 27.53 & 51.68  \\
        & MAAF \citep{MAAF} & - & 23.80 & 48.60 & 21.30 & 44.20 & 27.90 & 53.60  \\
        & CurlingNet \citep{CurlingNet} & - & 26.15 & 53.24 & 21.45 & 44.56 & 30.12 & 55.23  \\
        & RTIC-GCN \citep{RTIC} & - & 29.15 & 54.04 & 23.79 & 47.25 & 31.61 & 57.98  \\
        & CoSMo\citep{Lee_2021_CVPR_cosmo} & - & 25.64 & 50.30 & 24.90 & 49.18 & 29.21 & 57.46  \\
        & ARTEMIS\citep{ARTEMIS} & - & 27.16 & 52.40 & 21.78 & 43.64 & 29.20 & 53.83  \\
        & DCNet\citep{Kim_Yu_Kim_Kim_2021_dcnet} & - & 28.95 & 56.07 & 23.95 & 47.30 & 30.44 & 58.29 \\
        & SAC \citep{SAC} & - & 26.52 & 51.01 & 28.02 & 51.86 & 32.70 & 61.23  \\
        & FashionVLP\citep{FashionVLP} & - & 32.42 & 60.29 & 31.89 & 58.44 & 38.51 & 68.79  \\
        & LF-BLIP~\citep{cclip} & - & 25.31 & 44.05  &  25.39 & 43.57  &  26.54 & 44.48   \\
        & CASE~\citep{levy2023case} & \ding{51} & 47.44 & 69.36 & 48.48 & 70.23 & {50.18} & {72.24}  \\ 
        & CoVR-BLIP ~\citep{webvid-covr} & -   & 43.51 & 67.94 & 48.28 & 66.68 & 51.53 & 73.60  \\
        & CoVR-BLIP ~\citep{webvid-covr} & \ding{51} & 44.55 & 69.03 & 48.43 & 67.42 & 52.60 & 74.31  \\
        & Thawakar \textit{et.al} ~\citep{thawakar2024composed} & \ding{51} &  46.12 & 69.52& 49.61 & 68.88 & 53.79 & 74.74  \\
        \rowcolor{orange!15} \cellcolor{white}
        & \textbf{Our Approach} & \ding{51} & \textbf{48.12} & \textbf{71.48} & \textbf{51.38} & \textbf{70.38} & \textbf{55.08} & \textbf{75.96}  \\
        \midrule
        \multirow{5.2}{*}{\begin{tabular}[c]{@{}c@{}}  
        \rotatebox{90}{\small{Zero Shot}}\end{tabular}} 
        & Random & - & \textcolor{white}{0}0.26 & \textcolor{white}{0}1.31 & \textcolor{white}{0}0.16 & \textcolor{white}{0}0.79 & \textcolor{white}{0}0.19 & \textcolor{white}{0}0.95  \\
        & Pic2Word~\citep{saito2023pic2word} & \ding{51} & 20.00 & 40.20 & 26.20 & 43.60 & 27.90 & 47.40 \\
        & CoVR-BLIP ~\citep{webvid-covr} & \ding{51} &  21.95 & 39.05 & 30.37 & 46.12 & 30.78 & 48.73   \\
        & Thawakar \textit{et.al} ~\citep{thawakar2024composed} & - &  15.24 & 34.12 & 18.36 & 32.54 & 19.56 & 37.54  \\
        & \textbf{Our Approach} & \ding{51} &  \textbf{21.08} & \textbf{38.26} & \textbf{22.18} & \textbf{36.72} & \textbf{25.06} & \textbf{44.28}  \\
        & Thawakar \textit{et.al} ~\citep{thawakar2024composed} & \ding{51} &  24.57 & 40.93 & 33.12 & 48.42 & 33.16 & 50.24  \\
        \rowcolor{orange!15} \cellcolor{white}
        & \textbf{Our Approach} & \ding{51} &  \textbf{26.12} & \textbf{42.88} & \textbf{35.32} & \textbf{49.92} & \textbf{35.44} & \textbf{51.66}  \\
        \bottomrule
        \end{tabular}
        }
    \end{minipage}
    \vspace{-0.7em}
    \caption{
    \textbf{Left:} \textbf{Comparison of our approach with existing methods on the CIRR}~\citep{cirr} \textbf{test set}. We present the results in both training and zero-shot settings in terms of Recall@K and R@K. Our approach consistently outperforms existing methods, achieving Recall@K=1 score of 44.08 and Recall@K=50 score of 95.88. 
     \textbf{Right:} \textbf{Comparison with existing methods on retrieval tasks for specific attributes such as dress, shirt, and toptee on FashionIQ}~\citep{fashioniq} \textbf{validation set}. 
     Best results are in bold.
    }
    \label{tab:coir_methods_comparison}
    \vspace{-0.5em}

\end{table*}

\begin{table*}[t!]
    \centering
    \begin{minipage}{0.48\textwidth}
        \setlength{\tabcolsep}{17pt}
        \resizebox{\textwidth}{!}{%
        \begin{tabular}{l|cccc}
        \toprule
            Modification-Text & R@1 & R@5 & R@10 & R@50 \\
            \toprule
             WebVid-CoVR & 68.88 & 88.62 & 94.20 & 98.62 \\ 
             \textbf{Dense-WebVid-CoVR} & \textbf{71.26} & \textbf{89.12} & \textbf{94.56} & \textbf{98.88} \\ 
            \bottomrule
        \end{tabular}
        }
    \end{minipage}
    \hfill
    \begin{minipage}{0.48\textwidth}
        \resizebox{\textwidth}{!}{%
        \begin{tabular}{l|cccc}
        \toprule
            Our Model & R@1 & R@5 & R@10 & R@50 \\
            \toprule
            Without using dense descriptions in inference & 66.08 & 88.32 & 93.82 & 98.36 \\
            With using dense descriptions in inference & \textbf{71.26} & \textbf{89.12} & \textbf{94.56} & \textbf{98.88} \\ 
            \bottomrule
        \end{tabular}
        }
    \end{minipage}
    \vspace{-0.7em}
    \caption{
    \textbf{Left:} \textbf{Impact of using the modification text from the original WebVid-CoVR or from our Dense-WebVid-CoVR on the model performance at inference}.  
    Our method achieves best performance when incorporating dense and detailed modification texts from Dense-WebVid-CoVR dataset. Notably, our approach achieves an R@1 of 71.26, surpassing its performance of 68.88 with WebVid-CoVR. 
    \textbf{Right:} \textbf{Impact of using dense video descriptions during inference}. When utilizing dense descriptions, our method achieves superior performance across all metrics, achieving an R@1 of 71.26, compared to 66.08\% without using dense descriptions. Best results are in bold. 
    }
    \label{tab:ablation}
    \vspace{-1em}

\end{table*}

We further conduct experiments on the Ego-CVR dataset in a zero-shot setting. Tab.~\ref{tab:egocvr_comparison} shows the comparison on Ego-CVR test set. Here, recent methods such as TFR-CVR~\cite{hummel2024egocvr} uses the pre-trained TFR model that is trained on 10 million corpus of WebVid data for text-to-video retrieval. Our approach performs favorably in terms of global and local retrieval performance, compared to existing works. \\
\noindent \textbf{Ablation Study:} We first perform a study to analyze the impact of using dense modification text for CoVR. Tab.~\ref{tab:ablation} (left) compares the results of using our dense modification texts versus the WebVid-CoVR short modification texts. Our approach consistently outperforms across all recall metrics demonstrating the effectiveness of more detailed modification texts in improving the retrieval accuracy. We further examine the impact of dense descriptions during inference and present the results in Tab.~\ref{tab:ablation} (right). The model utilizing dense descriptions achieves a notable performance gain with a retrieval score of 71.26\% Recall@1, compared to 66.08\% without dense descriptions.

Table~\ref{tab:ablation_study2} shows that using Dense-WebVid-CoVR modification texts significantly improves retrieval accuracy, with Recall@1 increasing from 63.8 to 71.2 and Recall@5 from 87.5 to 89.1, compared to models trained with WebVid-CoVR modifications. This highlights that shorter modification texts often lead to incorrect target retrieval, causing models to focus on distractor videos similar to the input but lacking the intended changes. In contrast, richer modification texts help models capture subtle transformations more effectively, reinforcing the importance of detailed textual modifications in enhancing multimodal learning and preventing reliance on text-only retrieval.


We conduct a study to understand the impact of different fusion strategies on input video ($q$) and its description ($d$)
Here, we compared three fusion techniques to fuse visual embedding with description embedding: Addition, Cross-Attention (CA), and the proposed unified (weighted-mean). When using addition strategy in our framework, we achieve Recall@1 of 69.72. The results improve to 70.13 when using CA. The best results of Recall@1 of 71.26 are obtained when using the proposed unified fusion strategy. 

\subsection{Results on Composed Image Retrieval (CoIR)}

In addition to the CoVR task, we evaluate our approach on composed image retrieval (CoIR) task using both training and zero-shot settings. We conduct experiments on CIRR~\cite{cirr} test set and FashionIQ~\cite{fashioniq} validation set. The results are presented in Tab.~\ref{tab:coir_methods_comparison}.
On the left and when using the train CIRR setting, our method achieves  Recall@1 score of 56.30, outperforming recent approaches like CoVR-BLIP~\cite{webvid-covr} and Thawakar \textit{et al.}~\cite{thawakar2024composed} by a significant margin. 
In the zero-shot setting when the methods are not trained on CIRR data, our approach performs favorably against existing methods with a Recall@1 score of 44.08.
These results suggest the generalizability of our method in terms of handling unseen data.
We also analyze the performance on specific attribute-based retrieval tasks, as presented in Tab.~\ref{tab:coir_methods_comparison} (right). Our method achieves consistent improvements across various categories. Notably in the dress category, our method obtains R@50 score of 49.92. On the toptee category, our approach achieves  R@50 score of 51.66, outperforming existing methods in the zero-Shot setting. These results suggest that our method has the ability to handle attribute-specific composed image retrieval with better precision and accuracy.

\section{Conclusion}

We investigate the problem of composed video retrieval (CoVR) and propose a new dataset with detailed video descriptions and dense modification texts, capturing subtle visual and temporal changes. Our dataset, named Dense-WebVid-CoVR, comprises 1.6 million samples with dense modification text with an average length of 31.2 words. 
 
In addition, we propose an approach that encodes subtle changes by simultaneously processing input video, description, and dense modification text in a single grounding encoder. For CoVR, we conduct experiments on two datasets: Dense-WebVid-CoVR and Ego-CVR. Our approach achieves favorable results on both datasets. We further evaluate our approach for CoIR task on two datasets: CIRR and FashionIQ. Our approach achieves state-of-the-art performance on both datasets. A potential future direction is to explore CoVR task in a multilingual setting, especially for low-resource languages, to expand its real-world applicability to diverse populations. Another potential research direction is to explore efficient techniques for processing very long videos and their corresponding descriptions for the CoVR task.

\section{Acknowledgement}
The computations were enabled by resources provided by NAISS at Alvis partially funded by Swedish Research Council through grant agreement no. 2022-06725, LUMI hosted by CSC (Finland) and LUMI consortium, and by Berzelius resource provided by the Knut and Alice Wallenberg Foundation at the NSC.

{
    \small
    \bibliographystyle{ieeenat_fullname}
    \bibliography{main}
}

\clearpage
\maketitlesupplementary
\appendix

\section{Supplementary Materials}
In this appendix, we provide a detailed discussion of the related work on both Composed Image Retrieval (CoIR) and Composed Video Retrieval (CoVR). CoVR, as an extension of CoIR, brings the challenge of video temporal dynamics and detailed modifications into the retrieval space. We also present dataset statistics for our proposed fine-grained composed video retrieval dataset, highlighting the significant increase in the number of words in modification-text and the length of descriptions compared to the WebVid-CoVR dataset. In particular, our dataset provides substantially more detailed modification texts, averaging around 80 words per description, as compared to the sparse and shorter descriptions in WebVid-CoVR. This rich data allows for better handling of fine-grained video modifications, significantly improving retrieval performance.

\section{More on Related Work}

Composed image retrieval has evolved significantly with various approaches leveraging visual and textual inputs for accurate retrieval. Early methods like TIRG [Vo et al., 2019] employed hybrid embeddings but struggled with fine-grained details. Recent models, such as CLIP [Radford et al., 2021] and ARTEMIS [Delmas et al., 2022], improved retrieval by aligning images and text, though they often required large-scale pretraining. Cross-attention-based methods like CompoDiff [Gu et al., 2023] have shown promising results by better integrating visual and textual information. Our approach builds on these advancements by introducing dense modification texts, offering superior performance over state-of-the-art methods like BLIP [Li et al., 2023] and Thawakar et al. [2024], particularly in fine-grained and zero-shot retrieval tasks.

Several notable datasets have been developed to advance CoVR. EgoCVR~\cite{hummel2024egocvr}, one of the first benchmarks, emphasizes fine-grained action retrieval from egocentric videos, requiring temporal reasoning for accurate retrieval. WebVid-CoVR~\cite{webvid-covr} offers a large-scale dataset created automatically by mining video pairs with similar captions and generating modification texts using large language models, resulting in 1.6 million triplets. This dataset enables scalable, high-quality annotations for general-purpose CoVR tasks.
Recently, \cite{thawakar2024composed} proposes detailed language descriptions for source and target videos, improving context preservation and query-specific alignment. 
This work expands on WebVid-CoVR with detailed captions, further enhancing multi-modal embedding alignment and retrieval accuracy. 
In contrast, this work proposes a fine-grained CoVR dataset that offers dense and detailed modification texts, enabling models to capture subtle visual and temporal changes between videos. Unlike previous datasets, which often rely on sparse annotations, our approach provides comprehensive and context-rich data, making it a useful resource for training more accurate and robust retrieval models. The scalability and depth of our dataset push the boundaries of composed video retrieval, facilitating more precise video search across a wide range of real-world applications, from video editing to surveillance.

\section{Prompt used for Detailed Video Description Generation}
For generating detailed video descriptions, we utilized the WebVid-CoVR dataset, comprising diverse categories such as nature, lifestyle, and professional activities. The dataset’s videos, averaging around 16.8 seconds in length, offer substantial diversity in terms of visual content and actions. To effectively create captions for such videos, we employed Gemini-Pro~\cite{gemini}, a leading-edge video captioning model particularly suited for handling longer and more intricate videos. Unlike other models that often falter when processing extensive sequences, Gemini-Pro excels in maintaining a coherent narrative over time, making it ideal for this task. Its advanced temporal context-handling abilities ensure that even videos with numerous transitions and complex actions are described with high accuracy and detail.

\begin{figure}[t]
    \centering
    \includegraphics[width=1\linewidth]{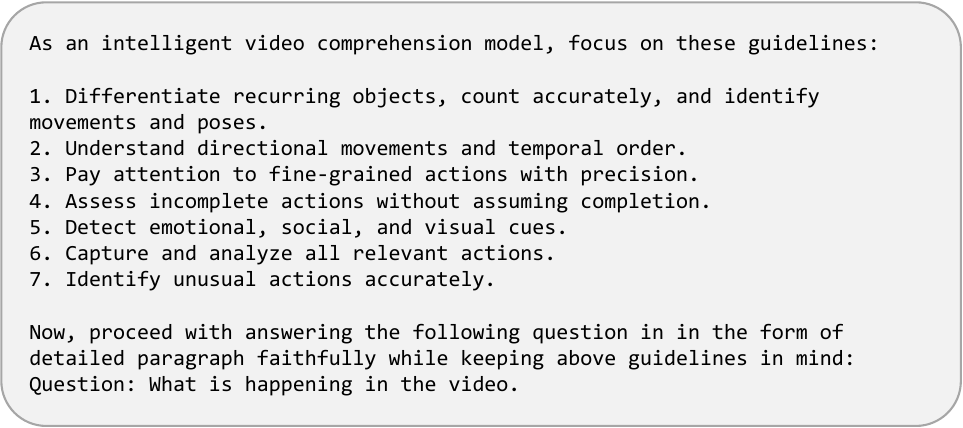}
    \caption{
    This figure presents an example prompt used to generate detailed video descriptions with the Gemini-1.5-Pro model. The prompt is designed to guide the model towards producing comprehensive, fine-grained descriptions by following specific guidelines. These guidelines instruct the model to accurately identify recurring objects, count them, capture directional movements, maintain temporal order, and assess actions with high precision. Additionally, the model is directed to detect emotional, social, and visual cues, ensuring a thorough understanding of the video's content. This structured prompt enables the model to deliver rich, context-aware descriptions that are essential for high-quality video retrieval tasks.
    }
    \label{fig:prompt_video_description}
\end{figure}

The video descriptions were generated using the prompt shown in Figure~\ref{fig:prompt_video_description}. This prompt was designed to extract comprehensive descriptions, ensuring the inclusion of all relevant details within the video. To ensure the quality and accuracy of the generated captions, we implemented a hallucination check using the BLIP model, which computes the cosine similarity between the input video and its corresponding caption. Video captions with a similarity score below 0.4 were considered inadequate and recomputed. This additional validation step ensures that the captions are both highly aligned with the video content and free from irrelevant information, resulting in a more robust and reliable dataset.

\section{Prompt used for detailed Modification Text Generation}
\begin{figure}[t!]
    \centering
    \includegraphics[width=1\linewidth]{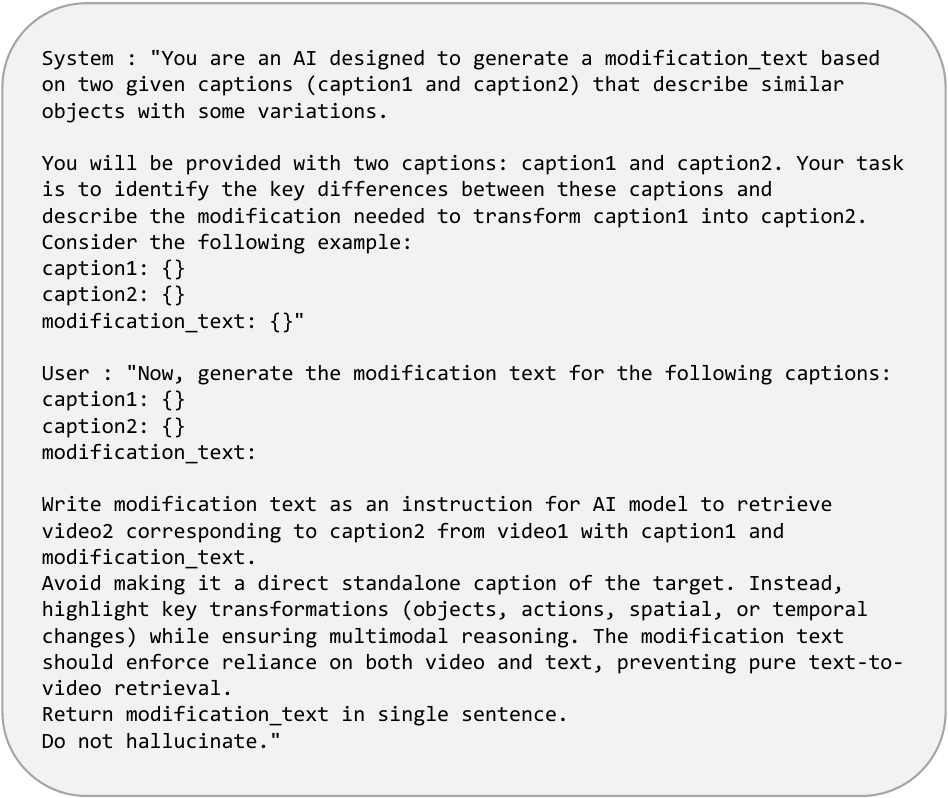}
    \caption{
    This figure illustrates an example prompt used with the GPT-4o model to generate precise modification texts between two video descriptions. The prompt instructs the model to analyze two given captions (caption1 and caption2), identify the key differences, and formulate a concise modification text to transform caption1 into caption2. The modification text is crafted as an instruction, guiding the retrieval model to locate the target video based on the described changes. By following these structured guidelines and avoiding any hallucinations, the prompt ensures that the generated modification text accurately reflects the required transformations, facilitating more precise and context-aware video retrieval.
    }
    \label{fig:prompt_modification_text}
\end{figure}

In fine-grained CoVR, modification text plays a key role in distinguishing two similar videos by clearly describing their differences, such as changes in actions, objects, or scenes. These texts focus on subtle variations to improve video retrieval based on fine details. To achieve this, we use GPT-4o~\cite{gpt4} to generate precise modification texts. The model is used with the help of in context learning using default triplets from WebVid, which include original video captions and their corresponding short modification texts, ensuring accuracy in generated detailed modification texts. The prompt shown in figure~\ref{fig:prompt_modification_text} was utilized to generate detailed modification texts. By leveraging this detailed prompt and using WebVid-CoVR triplets as a reference, GPT-4o was able to produce high-quality, contextually rich modification texts.

\section{Quality Control Protocol for Human Verification of the generated triplets}
\label{app:sec-verification}
\begin{figure*}[ht]
    \centering
    \includegraphics[width=1\textwidth]{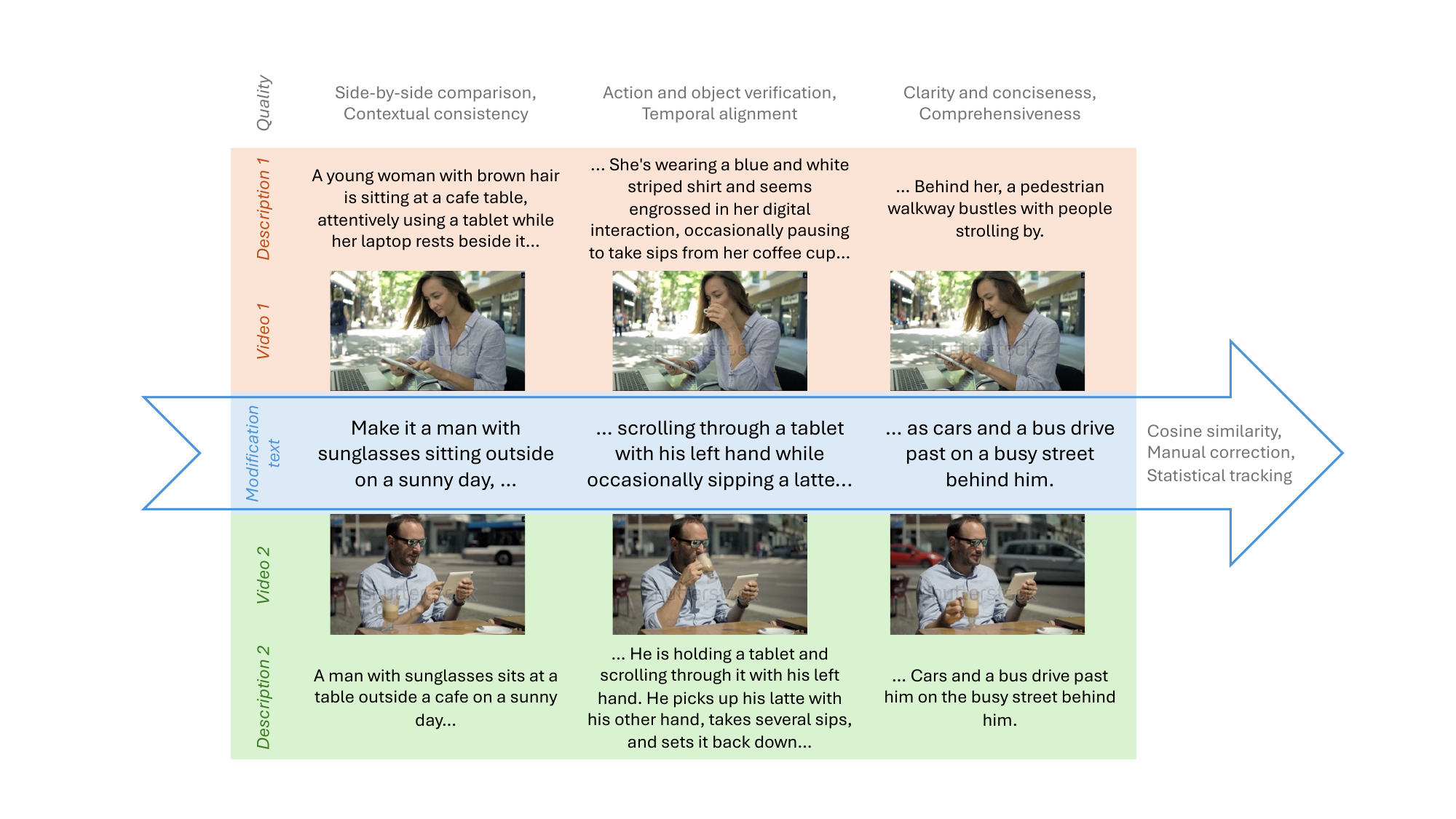}
    \caption{A data verification pipeline used for quality assessment of the generated data in the evaluation set. By ensuring a specific set of desired qualitative properties (in grey) of the textual data, we achieve high-quality video descriptions (in orange and green) and modification texts (in blue).
    }
    \label{fig:verif_pipeline}
\end{figure*}

The manual human verification process involves trained annotators who systematically validate and refine the generated modification texts. Annotators are presented with side-by-side input and target videos, along with their corresponding modification text, and are responsible for assessing the correctness and completeness of the descriptions. To enhance inter-annotator reliability, we employ a consensus-based approach, where a subset of the dataset undergoes multiple rounds of validation.
In order to assess the quality of the generated data and increase reliability, we provide a human-verified evaluation set, where each triplet is manually checked. Specifically, the overall verification pipeline and qualitative properties of the generated data are depicted in Figure ~\ref{fig:verif_pipeline}. To ensure the correctness of both automated and manual modifications, we employ the following multi-check verification protocol:

\begin{itemize}
   \item \textbf{Side-by-Side Comparison:} Present the input and target videos alongside their corresponding modification text to the reviewers. Ensure that the modifications mentioned accurately reflect the differences between the two videos.
   \item \textbf{Contextual Consistency Check:} Verify that the modification text addresses 
   key changes in the visual or action-based context, such as consistent object movement and transitions in the main scene, related surroundings, and background.
    key changes in the visual or action-based context, such as object movement, color changes, scene transitions, or actions.
   \item \textbf{Action and Object Verification:} Reviewers should check that the objects and their corresponding actions mentioned in the modification text are present in both the input and target videos and that the text accurately describes their differences.
    \item \textbf{Temporal Alignment:} Ensure that the modification text aligns with the actual sequence of actions or events in the videos. The text should reflect changes in timing or flow between the two videos (e.g., one action follows another).
   \item \textbf{Comprehensiveness of Description:} The modification text should cover all relevant changes between the input and target video, providing a clear and complete description of the differences without leaving out important details.
   \item \textbf{Clarity and Conciseness:} Ensure that the modification text is clear, concise, and free of ambiguity. It should effectively communicate the modification in a straightforward and readable manner.
   \item \textbf{Cosine Similarity Threshold:} Use a cosine similarity threshold (e.g., 0.4) to identify low-quality modification texts that might require further human review and adjustment.
    \item \textbf{Statistical Tracking:} Maintain records of 
    how many triplets were verified, how many corrections required, and what types of errors were most common. This helps refine the verification process over time.
   the number of verified triplets, required corrections, and types of common errors to refine the verification process over time.
    \item \textbf{Manual Corrections:} Reviewers can make manual corrections to the modification text if there are errors, missing details, or irrelevant information. 
    They should focus on improving the alignment between text and video content.
\end{itemize}

\noindent\textbf{Impact of Training Set Errors: } Given the scale of the training set, it is reasonable to expect some inherent noise in the unverified portion. However, our multi-stage quality control ensures that errors are minimized and do not significantly impact the model’s performance. Empirically, we observe that training on denser, high-quality modification texts leads to significant improvements in retrieval accuracy (+3.4\% Recall@1, as shown in Table 2 of the main paper). Furthermore, experiments on the manually verified subset demonstrate that even partial high-quality annotations substantially enhance the model’s generalization ability.

In summary, our dataset combines large-scale automated generation with meticulous human verification, ensuring that the test set is fully validated and the training set maintains a high standard of correctness. This makes Dense-WebVid-CoVR a robust and reliable benchmark for fine-grained composed video retrieval.

\section{Dataset Statistics}
\label{app:sec-statistics}

\begin{figure*}[ht]
    \centering
    \begin{subfigure}{0.49\textwidth}
        \centering
        \includegraphics[width=\linewidth]{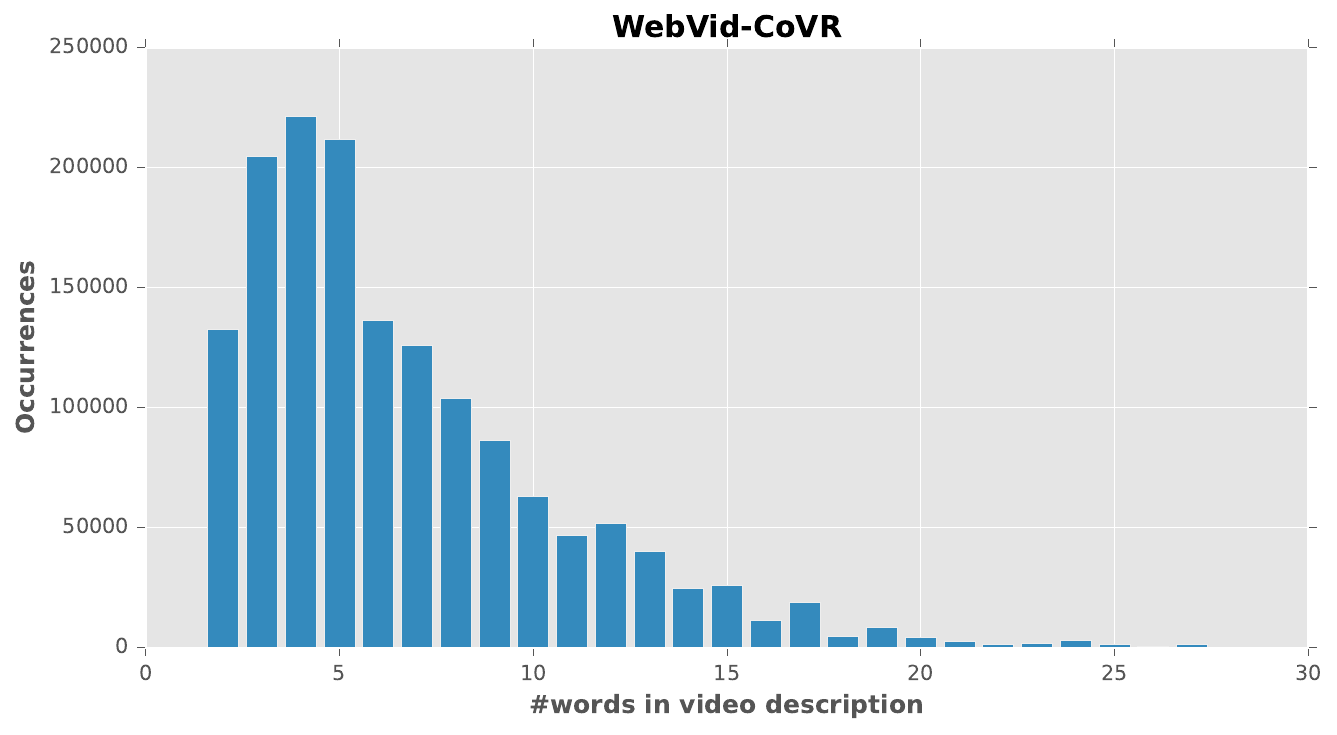} 
    \end{subfigure}
    \hfill
    \begin{subfigure}{0.49\textwidth}
        \centering
        \includegraphics[width=\linewidth]{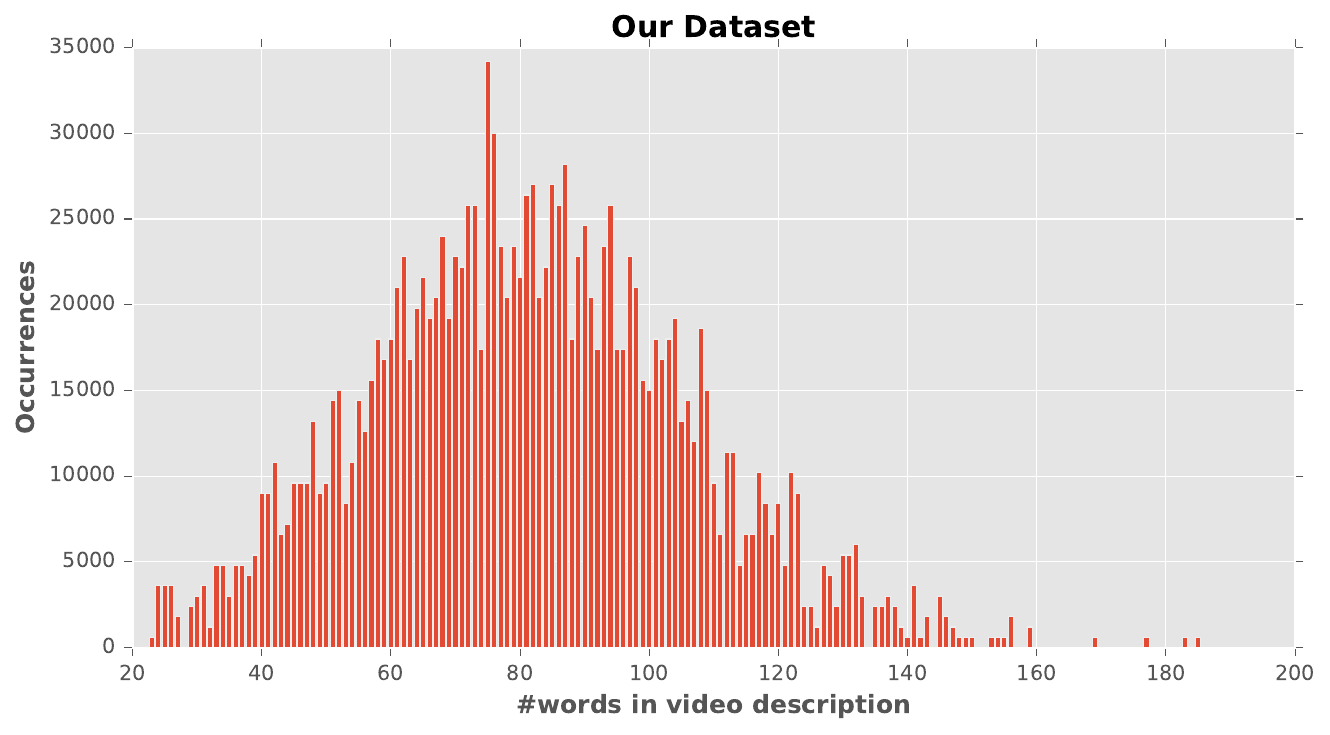} 
    \end{subfigure}
    \caption{
    The comparison of the distribution of the number of words in video descriptions between the WebVid-CoVR~\citep{webvid-covr} dataset (left) and our proposed dataset (right). 
    In the WebVid-CoVR dataset, the majority of video descriptions are short, typically containing between 3 and 10 words, with a sharp decline beyond that. 
    This limited description length often lacks the depth required for fine-grained video retrieval tasks. 
    In contrast, our dataset features significantly longer and more detailed video descriptions, with the distribution centered around 80 to 100 words per description. 
    The increased richness and granularity of these descriptions enable more precise alignment between video content and modification text, greatly enhancing retrieval performance. This detailed annotation provides a stronger foundation for training retrieval models that can handle complex, fine-grained modifications in video content.
    }
    \label{fig:description-count-compare}
\end{figure*}

As extra statistical analysis, in Fig. \ref{fig:description-count-compare} we compare the distribution of the number of words in video descriptions between the WebVid-CoVR~\citep{webvid-covr} dataset and our proposed Dense-WebVid-CoVR dataset. In the WebVid-CoVR dataset, the majority of video descriptions are short, typically containing between 3 and 10 words, with a sharp decline beyond that. This limited description length often lacks the depth required for fine-grained video retrieval tasks. In contrast, our Dense-WebVid-CoVR dataset features significantly longer and more detailed video descriptions, with the distribution centered around 80 to 100 words per description. The increased richness and granularity of these descriptions enable more precise alignment between video content and modification text, greatly enhancing retrieval performance. This detailed annotation provides a stronger foundation for training retrieval models that can handle complex, fine-grained modifications in video content.

\begin{figure}[t]
    \centering
    \includegraphics[width=1\linewidth]{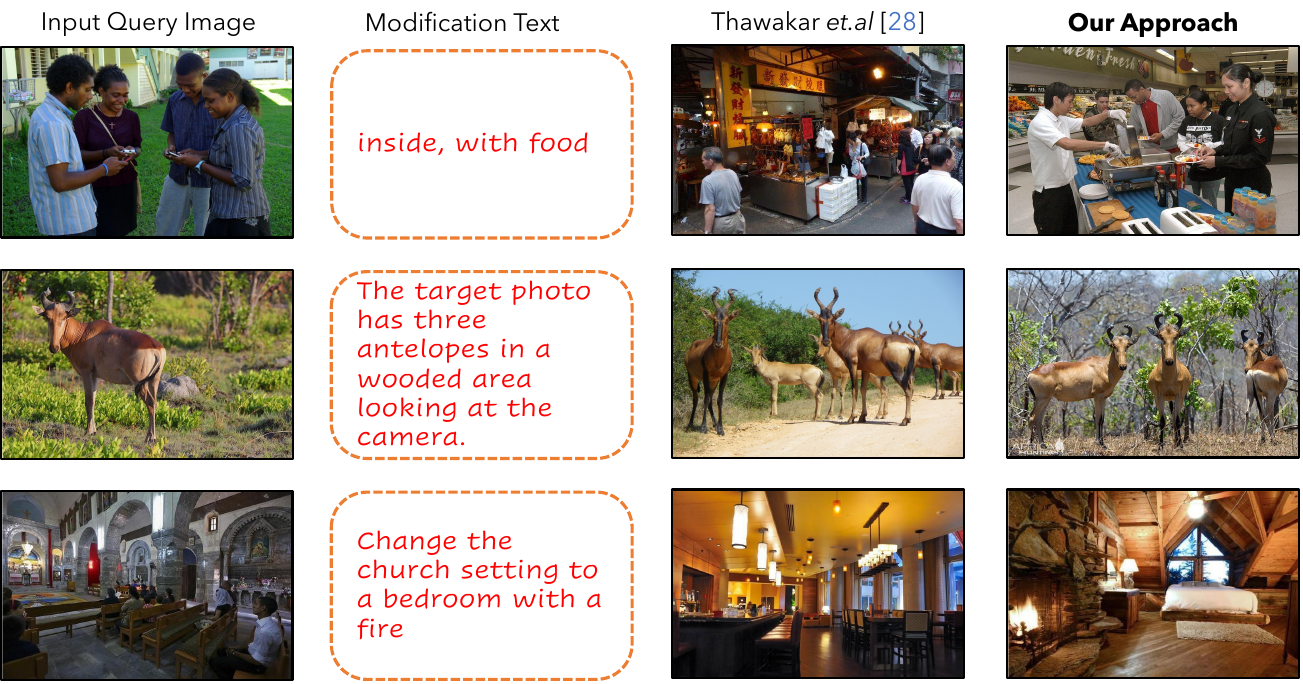}
    \vspace{-1.5em}
    \caption{
    Qualitative comparison between the recent CoVR model~\cite{thawakar2024composed}) and our approach for zero-shot composed image retrieval (CoIR) task on CIRR validation set. Our unified fusion method captures the modification text more effectively, resulting in more accurate retrievals that align closely with the intended changes in the input query image. Best viewed zoomed in. Additional results are presented in suppl. material.
    }
    \label{fig:qual_comparison_cirr}
    \vspace{-1em}
\end{figure}

\begin{figure*}[t!]
    \centering
    \begin{subfigure}{0.49\textwidth}
        \centering
        \includegraphics[width=\linewidth]{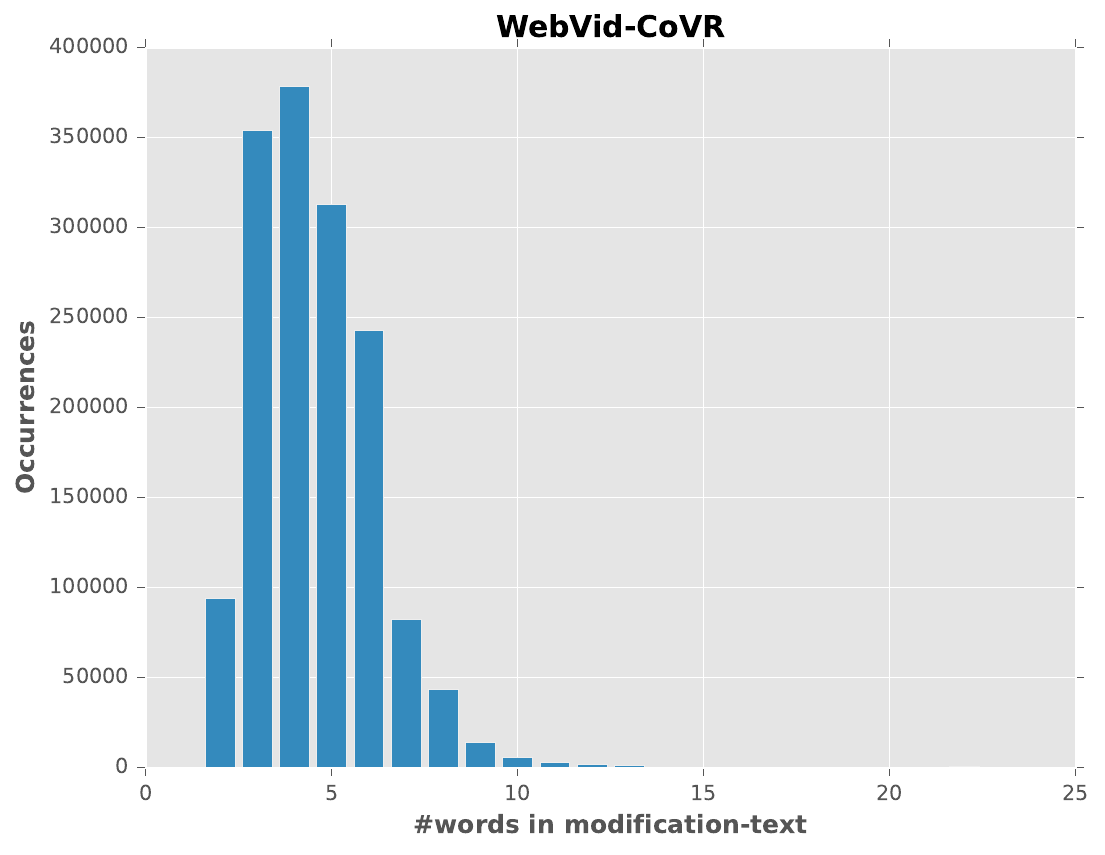} 
    \end{subfigure}
    \hfill
    \begin{subfigure}{0.49\textwidth}
        \centering
        \includegraphics[width=\linewidth]{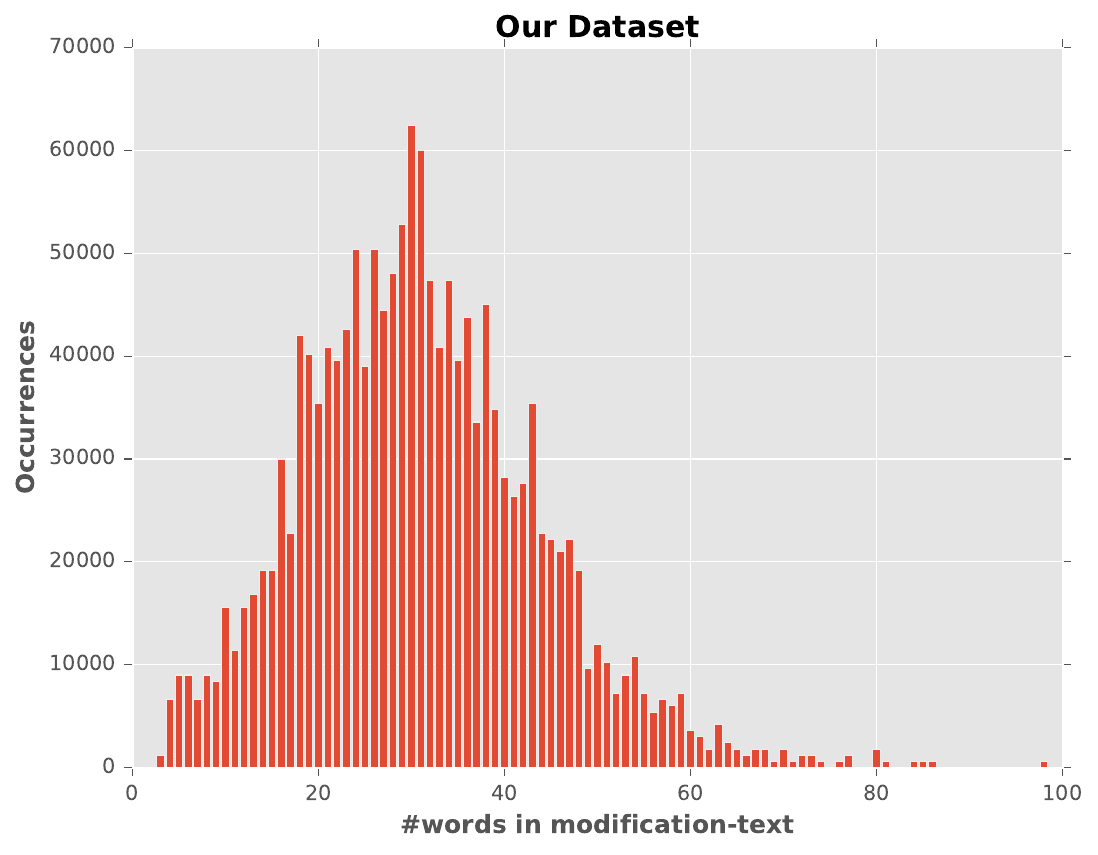} 
    \end{subfigure}
    \caption{The graphs illustrate the distribution of word counts in modification texts between the WebVid-CoVR~\citep{webvid-covr} dataset (left) and our proposed dataset (right). 
    In the WebVid-CoVR dataset, the majority of modification texts are very brief, with a word count predominantly ranging between 3 and 7 words. 
    This brevity can limit the model’s ability to capture the distinct modifications needed for fine-grained video retrieval. 
    In contrast, our dataset provides significantly more detailed modification texts, with the distribution centered around 30 to 50 words. 
    The increased length and richness of our modification texts allow for a more comprehensive representation of the desired changes between videos, facilitating improved retrieval accuracy. By offering such detailed descriptions, our dataset enables models to handle complex modifications more effectively, leading to superior performance in fine-grained composed video retrieval tasks.
    }
    \label{fig:modification-text-count-compare}
\end{figure*}

Additionally, Figure \ref{fig:modification-text-count-compare} illustrates the distribution of word counts in modification texts between the WebVid-CoVR~\citep{webvid-covr} dataset and our proposed dataset. We can observe that in the WebVid-CoVR dataset, the majority of modification texts are very brief, with a word count predominantly ranging between 3 and 7 words. This brevity can limit the model’s ability to capture the distinct modifications needed for fine-grained video retrieval. In contrast, our dataset provides significantly more detailed modification texts, with the distribution centered around 30 to 50 words. The increased length and richness of our modification texts allow for a more comprehensive representation of the desired changes between videos, facilitating improved retrieval accuracy. By offering such detailed descriptions, our dataset enables models to handle complex modifications more effectively, leading to superior performance in fine-grained composed video retrieval tasks.

\section{Qualitative Analysis}

\begin{figure*}[ht]
    \centering
    \includegraphics[width=0.99\textwidth]{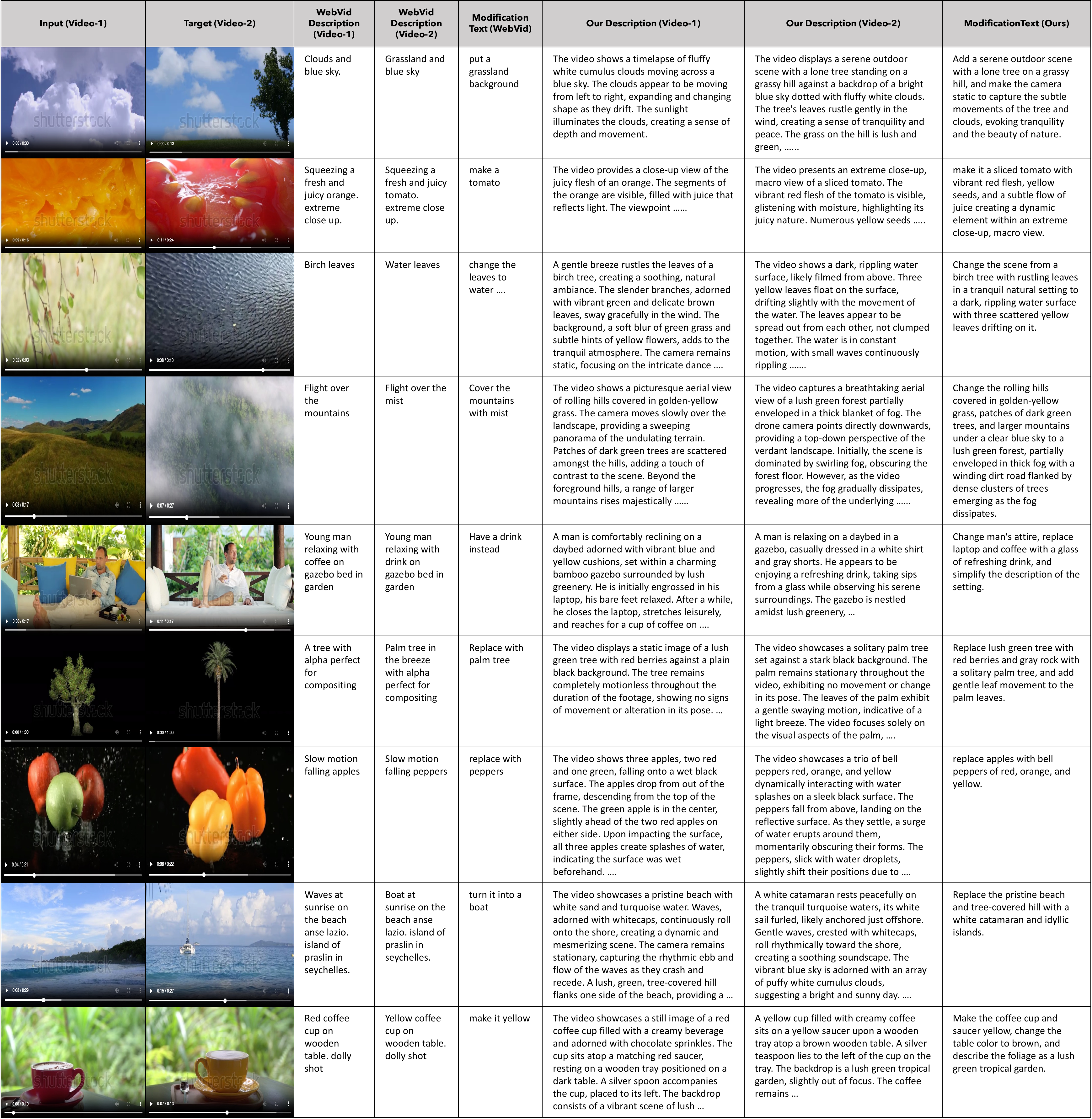}
    \caption{Data samples comparison from WebVid-CoVR our improved Dense-WebVid-CoVR dataset.}
    \label{fig:dataset_samples}
\end{figure*}

 Figure~\ref{fig:dataset_compare} illustrates the impact of fine-grained modification texts by comparing the input query video and target retrieval video. It highlights our fine-grained modification text's ability to capture subtle textual differences and retrieve the desired videos accurately. Figure~\ref{fig:dataset_samples} presents a comparison of datasets used in our study, emphasizing the diversity and richness of video-caption pairs in our proposed dataset, which contributes to the effective fine-grained retrieval performance. Figure ~\ref{fig:qual_comparison} provides qualitative comparisons, showcasing retrieval outputs for various queries between Thawakar \textit{et.al}~\cite{thawakar2024composed} and our proposed approach on Dense-WebVid-CoVR set. These examples demonstrate that our proposed dataset and approach consistently aligns with the specified modifications, outperforming baseline methods in understanding and applying detailed textual inputs. This supplementary material further validates the robustness and reliability of our proposed method in fine-grained video retrieval.

 Figure.~\ref{fig:qual_comparison_cirr} presents a qualitative between~\cite{thawakar2024composed} and our approach on examples from CIRR validation set in the zero-shot setting. Compared to~\cite{thawakar2024composed}, our approach achieves favorable retrieval performance. 
For instance, the modification text in the second row is: \textit{The target photo has three antelopes in a wooded area looking at camera}. Our approach provides accurate retrieval with respect to object count (three antelopes) and background context (wooded area). 
Our approach achieves accurate retrieval results with respect to the object count (three antelopes) and the background concept (wooded area).

\begin{figure*}[ht]
    \centering
    \includegraphics[width=0.8\textwidth]{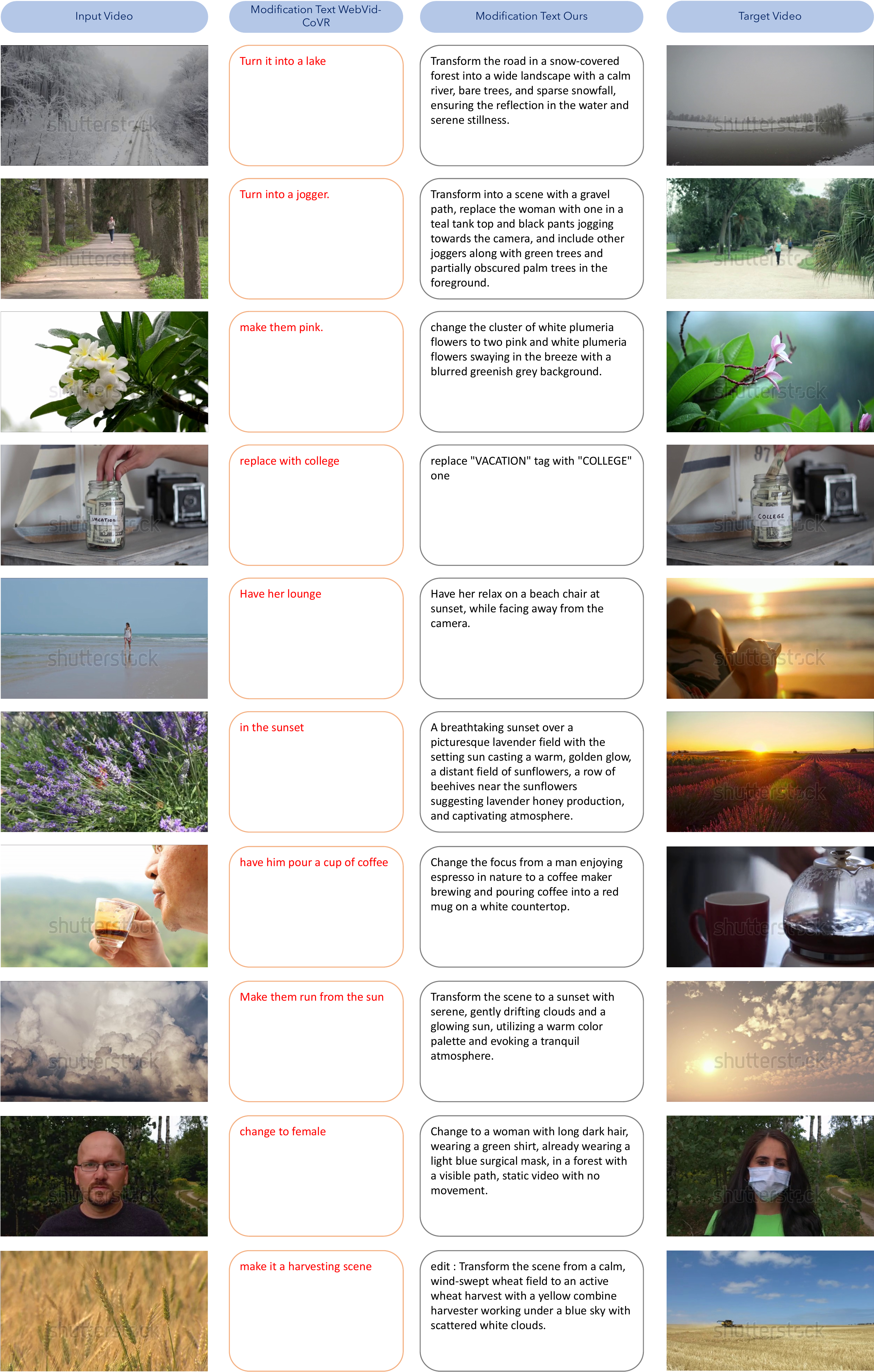}
    \caption{The modification-text comparison between WebVid-CoVR and our improved Dense-WebVid-CoVR dataset.}
    \label{fig:dataset_compare}
\end{figure*}

\begin{figure*}[t!]
    \centering
    \includegraphics[width=0.99\textwidth]{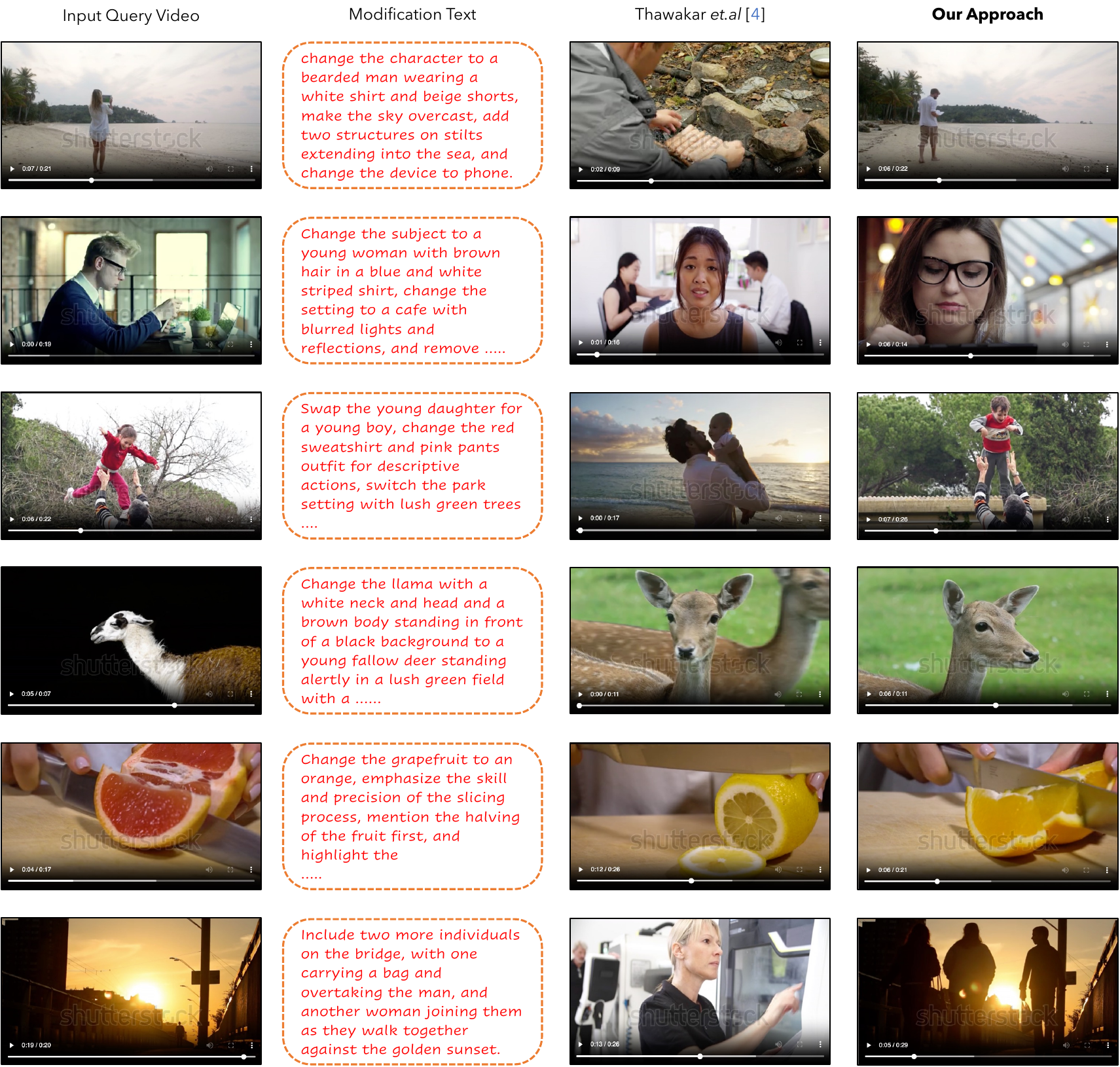}
    \caption{
    The figure demonstrates the effectiveness of fine-grained modification text in video retrieval, comparing the proposed approach with Thawakar \textit{et al.}~\cite{thawakar2024composed}. Each row showcases a query video, a detailed modification text specifying changes in subjects, actions, objects, or scenes, and the results retrieved by both methods. The proposed approach consistently retrieves videos that accurately reflect the described modifications, such as changes in characters, settings, objects, or activities, while the comparison method often fails to adapt to the nuanced details. This highlights the proposed method's superior ability to interpret and apply fine-grained textual instructions for precise video retrieval.
    }
    \label{fig:qual_comparison}
\end{figure*}

\end{document}